\documentclass{article}

\usepackage{wrapfig}
\usepackage{microtype}
\usepackage{multirow}
\usepackage{graphicx}
\usepackage{subfigure}
\usepackage{pifont}
\usepackage{transparent}
\usepackage{booktabs} 
\usepackage{threeparttable}
\usepackage{hyperref}
\usepackage{longtable}
\usepackage{threeparttablex}

\newcommand{\boldres}[1]{{\textbf{\textcolor{red}{#1}}}}
\newcommand{\secondres}[1]{{\underline{\textcolor{blue}{#1}}}}

\usepackage[accepted]{icml2025}

\usepackage{amsmath}
\usepackage{amssymb}
\usepackage{mathtools}
\usepackage{amsthm}

\usepackage[capitalize,noabbrev]{cleveref}

\theoremstyle{plain}

\theoremstyle{definition}

\theoremstyle{remark}

\usepackage[textsize=tiny]{todonotes}

\icmltitlerunning{TimesBERT: A BERT-Style Foundation Model for Time Series Understanding}

\begin{document}

\twocolumn[
\icmltitle{TimesBERT: A BERT-Style Foundation Model for Time Series Understanding}



\icmlsetsymbol{equal}{*}

\begin{icmlauthorlist}
\icmlauthor{Haoran Zhang}{equal,software}
\icmlauthor{Yong Liu}{equal,software}
\icmlauthor{Yunzhong Qiu}{equal,software}
\icmlauthor{Haixuan Liu}{software}
\\
\icmlauthor{Zhongyi Pei}{software}
\icmlauthor{Jianmin Wang}{software}
\icmlauthor{Mingsheng Long}{software}
\end{icmlauthorlist}

\icmlaffiliation{software}{
School of Software, BNRist, Tsinghua University. Haoran Zhang $<$zhang-hr24@mails.tsinghua.edu.cn$>$. 
Yong Liu $<$liuyong21@mails.tsinghua.edu.cn$>$.  
Yunzhong Qiu $<$qiuyz24@mails.tsinghua.edu.cn$>$}
\icmlcorrespondingauthor{Mingsheng Long}{mingsheng@tsinghua.edu.cn}

\icmlkeywords{Large time series models, Data-centric AI, Self-supervised pre-training, Transformers}

\vskip 0.3in
]



\printAffiliationsAndNotice{\icmlEqualContribution} 

\begin{abstract}
Time series analysis is crucial in diverse scenarios. Beyond forecasting, considerable real-world tasks are categorized into classification, imputation, and anomaly detection, underscoring different capabilities termed \emph{time series understanding} in this paper. While GPT-style models have been positioned as foundation models for time series forecasting, the BERT-style architecture, which has made significant advances in natural language understanding, has not been fully unlocked for time series understanding, possibly attributed to the undesirable dropout of essential elements of BERT. In this paper, inspired by the shared multi-granularity structure between multivariate time series and multisentence documents, we design TimesBERT to learn generic representations of time series including temporal patterns and variate-centric characteristics. In addition to a natural adaptation of masked modeling, we propose a parallel task of \emph{functional token prediction} to embody vital multi-granularity structures. Our model is pre-trained on 260 billion time points across diverse domains. Leveraging multi-granularity representations, TimesBERT achieves state-of-the-art performance across four typical downstream understanding tasks, outperforming task-specific models and language pre-trained backbones, positioning it as a versatile foundation model for time series understanding.
\end{abstract}

\section{Introduction}

\begin{figure}[t]
\begin{center}
    \centerline{\includegraphics[width=.98\columnwidth]{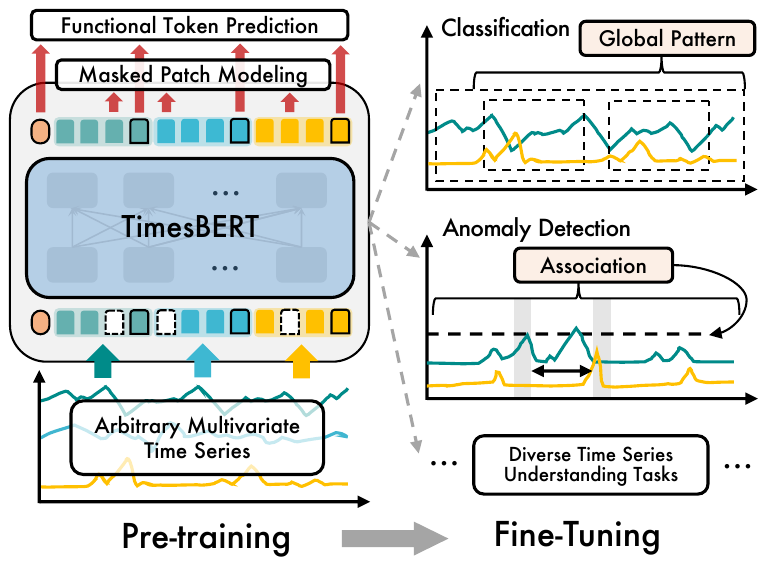}}
    \vspace{-5pt}
	\caption{TimesBERT inherits and extends the pre-training and fine-tuning paradigm established by BERT, which learns generalizable representation through pre-training on large-scale datasets of arbitrary multivariate time series, and adapts the foundation model to diverse tasks of time series understanding.} 
	\label{fig:intro}
\end{center}
\vspace{-30pt}
\end{figure}

Time series analysis is extensively applied across numerous practical applications and has a diverse form of tasks, among which time series forecasting has attracted significant attention and research efforts~\cite{oreshkin2019n, wu2021autoformer, nie2022time, liu2023itransformer}, becoming a primary task for evaluating the advances of deep learning methods. However, the remaining tasks have received relatively limited focus, resulting in a lack of comprehensive explorations on the model capabilities for practical demands. As shown in Figure~\ref{fig:intro}, for tasks such as time series classification~\cite{Franceschi2019UnsupervisedSR} and anomaly detection~\cite{xu2021anomaly}, multifaceted patterns in the context, such as bidirectional temporal dependencies, variate-centric representations, and mutual correlations between multiple variates, can outweigh causal dependencies and local variations emphasized in forecasting.  It underscores a generic representation learning capability from multi-granularity structures in multivariate time series. We collectively refer to this paradigm as \emph{time series understanding}.

\begin{figure*}[ht]
\begin{center}
    \center{\includegraphics[width=\textwidth]{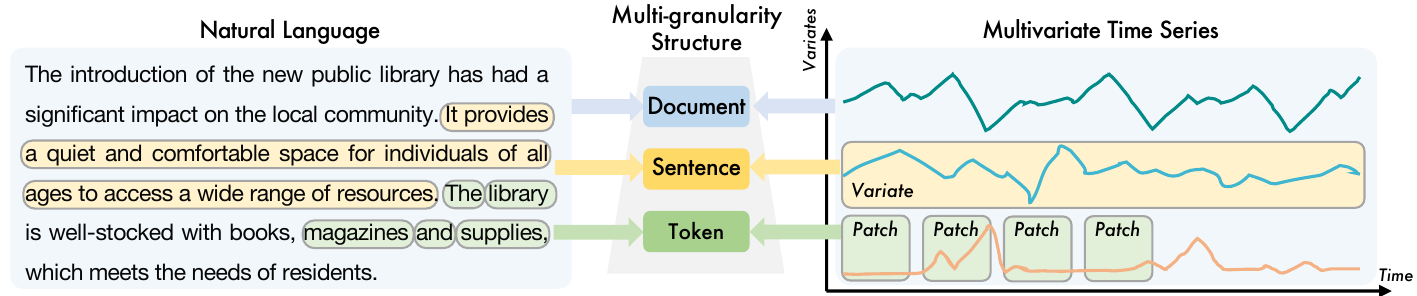}}
    \vspace{-15pt}
	\caption{\textbf{A multivariate time series is worth a natural language document.} We propose to fully repurpose BERT for learning structured representations of time series. The representations embodying different granularities can facilitate diverse time series understanding tasks.} 
	\label{fig:data}
\end{center}
\vspace{-10pt}
\end{figure*}

Foundation models~\cite{radford2018improving, dosovitskiy2020image} have advanced significantly in generalization performance, making them a promising solution for data-scarce and task-agnostic applications. While prevailing GPT-style models excel in generative tasks like time series forecasting ~\cite{das2023decoder, liu2024timer}, they lack the ability to leverage bidirectional context, causing a critical bottleneck for global understanding. By contrast, BERT~\cite{devlin2018bert} has exhibited task-versatility in natural language understanding like sentiment classification and entity identification. In addition to the primary objective of masked language modeling (MLM),  BERT facilitates functional tokens \texttt{[CLS]} and \texttt{[SEP]} to enable modeling multi-granularity structures from multisentence text documents, in compliance with the auxiliary task of next-sentence prediction (NSP) to reason about sentence-wise relationships. 
Moreover, BERT has popularized the pre-training and fine-tuning paradigm, which can facilitate a broader range of distinct downstream tasks, which has however not been unleashed in large-scale pre-trained time series foundation models.

As illustrated in Figure~\ref{fig:data}, time series share surprising structural similarities with natural language. In addition to the prevalent practice of regarding a patch as a token \cite{nie2022time}, we observe that a sufficiently long context of time series contains rich semantics to reveal variate-wise characteristics of time series, which possesses a similar structural correspondence with a sentence in natural language. On this basis, we highlight that  ``\emph{a multivariate time series is worth a multisentence text document},'' allowing us to inherit BERT, a general-purpose language representation learning framework with remarkable contextual awareness and multi-granularity capabilities for document-like data structures, to extract generic representations from heterogeneousness time series. The outcome pre-trained model manifested as a general feature extractor of time series, can facilitate a variety of understanding tasks. Nevertheless, while BERT adopts a fixed number of sentences with a fixed length for NSP, multivariate series can vary in both temporal length and variate number. It is required to implement a unified embedding on BERT, which is to adhere to large-scale pre-training and adaptation to various understanding tasks.

Inspired by the above motivations, we propose TimesBERT, a pre-trained foundation model for time series understanding. We conduct large-scale pre-training on 260 billion time points collected from multiple domains. We devise a unified embedding mechanism and repurpose the functional tokens in BERT to align with multivariate time series. 
In contrast to the previous methods that prevalently employed Channel-Independence~\cite{nie2022time}, in accordance with our established framework for interpreting multivariate time series as analogous to documents, we implement the pre-training of any-variate and any-length time series to handle the discrepancies in variate- and sentence-wise modeling, which reserves inherently structured representations of time series such as variate correlation. TimesBERT achieves significant improvement on four typical understanding tasks and  113 real-world datasets compared with state-of-the-art task-specifc models, exhibiting outstanding transferability.

Our contributions can be summarized as follows:
\begin{itemize}
    \item We rethink the common appeal of representation learning for time series understanding, and propose to treat multivariate time series as multisentence documents, revealing the advantages of BERT as a pre-trained model.
    \item We develop TimesBERT, which consists of a unified structured embedding and a functional token prediction task toward the multi-granularity structure of multivariate time series, fully aligning BERT to time series.
    \item We pre-train our model on large-scale dataset with 260 billion time points, which can be adapted with state-of-the-art results on time series classification, imputation, anomaly detection,  and short-term forecasting tasks.
\end{itemize}
\vspace{-10pt}

\section{Related Works}

\begin{figure*}[t]
\begin{center}
    \center{\includegraphics[width=\textwidth]{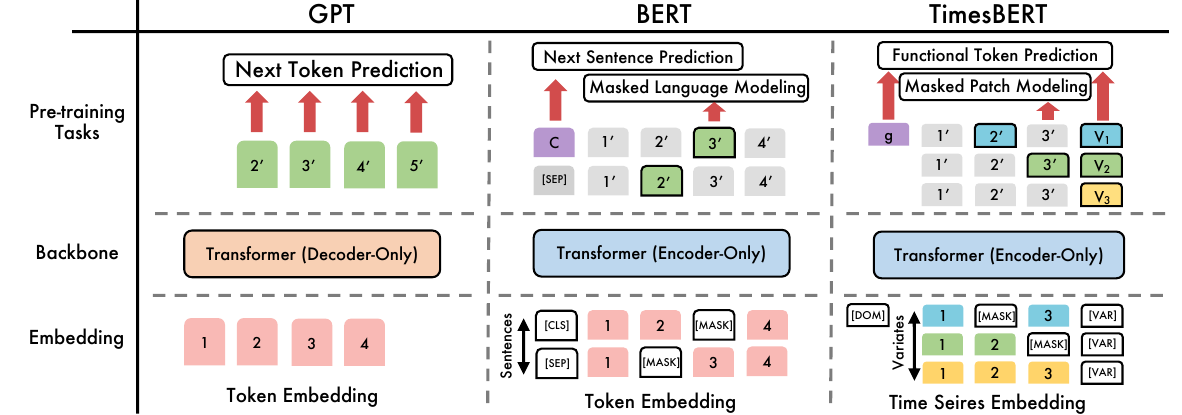}}
    \vspace{-15pt}
	\caption{Comparison between GPT~\cite{radford2018improving}, BERT~\cite{devlin2018bert}, and TimesBERT on embedding, backbone, and training objective. In contrast to BERT's sentence pair formulation, we implement an embedding approach for data with an arbitrary number of variates and design corresponding functional tokens to accommodate the inherent irregularity of time series variates.} 
	\label{fig:comparation}
\end{center}
\vspace{-10pt}
\end{figure*}

\subsection{Time Series Understanding}

Time series understanding includes a series of tasks that require structured representations and semantic extraction. 
Classical time series understanding methods such as Dynamic Time Warping \cite{Berndt1994UsingDT} and Isolation Forests \cite{bandaragoda2018isolation} make use of statistical-based representations to identify temporal motifs. 
Subsequent works~\cite{wu2022timesnet,donghao2024moderntcn} based on CNN backbones preliminarily exhibit the ability of deep learning-based models in time series understanding. 
Prevailing Transformer-based models \cite{zhou2021informer,wu2021autoformer,nie2022time,liu2023itransformer} apply attention mechanisms to discover potential correlations among different granularities. However, most deep learning models are originally designed for forecasting tasks with insufficient adaptation for time series understanding.

\subsection{BERT-Style Models}
Developed for natural language processing, BERT~\cite{devlin2018bert} conducted pioneer work in highlighting the significance of bidirectional information for data comprehension and demonstrates the effectiveness of the pre-training fine-tuning paradigm. More significantly, BERT introduces a structured and generic view to analyzing words, sentences, and documents, as opposed to simply considering natural language as entirely serialized entities.

While subsequent models extended capabilities of BERT for natural language~\cite{liu2019roberta, lan2019albert}, BERT-style models also exhibit wide-ranging effectiveness in other data modalities. MAE~\cite{he2022masked} employs an asymmetric encoder-decoder structure within the framework of masked modeling, achieving substantial pre-training improvements in image classification tasks. BEiT~\cite{bao2021beit} employs VQVAE~\cite{van2017neural} to convert images into corresponding discrete semantic representations. Notably, models like T5~\cite{raffel2020exploring} and GPT-3~\cite{brown2020language} are the counterpart of BERT-style architecture, with encoder-decoder T5 unifying tasks into a text-to-text framework and decoder-only GPT-3 leveraging massive scale for generative modeling, pushing the limits of scalable models.

\subsection{Pre-trained Time Series Models}

Pre-training methods in the field of time series have achieved advancements in building task-specific and foundation models. TST~\cite{zerveas2021transformer} and PatchTST~\cite{nie2022time} employ BERT-style masked pre-training at the point level and patch level, respectively. SimMTM~\cite{dong2023simmtm} attempts to integrate neighbor data comparison with masked point modeling. Free from respectively fine-tuning,  TimesFM~\cite{das2023decoder}, Timer~\cite{liu2024timer, liu2024timerxl}, and Chronos~\cite{ansari2024chronos} exhibit advantages of zero-shot forecasting through large-scale pre-training. However, these models primarily focus on forecasting-based tasks, lacking task versatility for distinct understanding tasks.

There have been several initial explorations on BERT-style pre-trained models. MOMENT~\cite{goswami2024moment} utilizes the T5 encoder for pre-training to achieve downstream multi-task capabilities. 
Moirai~\cite{woo2024unified} achieves multivariate embedding and employs masked modeling by forecasting the future patches. VisionTS~\cite{chen2024visionts} exhibits the robust transferability of vision-masked autoencoders across different modalities. However, essential elements for structured representation learning in BERT are not fully leveraged. Therefore, we delve into aligning time series with multisentence documents and next sentence prediction tasks, thus innovatively repurposing TimesBERT's pre-training objective to present a versatile pre-trained model for diverse understanding tasks.

\section{Approach}

TimesBERT employs an encoder-only Transformer to learn structured representations of multivariate time series, which is aligned with BERT~\cite{devlin2018bert} in both model architecture and objective design as shown in Figure~\ref{fig:comparation}. By pre-training on 260 billion time points from different domains, we present a task-versatile foundation model, which can be fine-tuned for various time series understanding tasks.

\begin{figure*}[ht]
\begin{center}
    \center{\includegraphics[width=\textwidth]{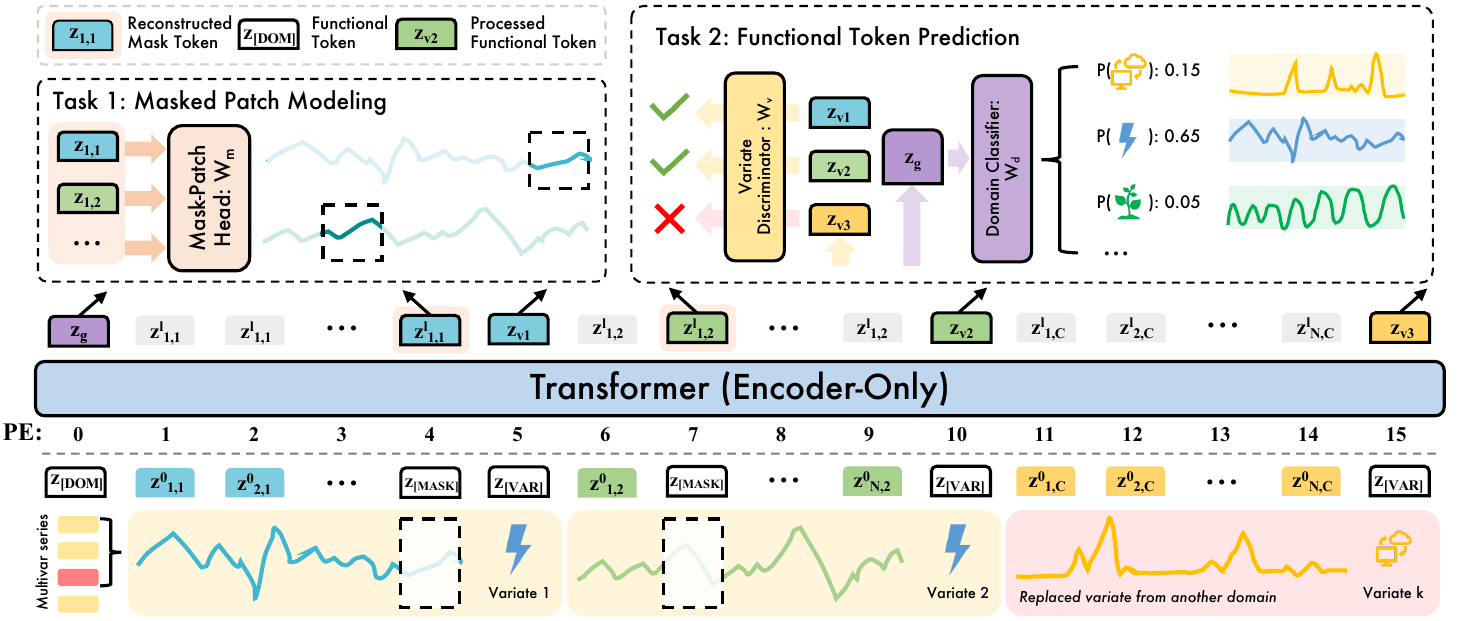}}
    \vspace{-15pt}
	\caption{Illustration of the TimesBERT architecture and pre-training objectives. 
    The input multivariate time series is embedded into a token sequence for the transformer encoder following a unified time series embedding process that includes patching, function token insertion, and flattening. Following the output from the backbone, the reconstructed patches and functional tokens are respectively fed into corresponding pre-training tasks including MPM and FTP, ultimately forming the joint optimization objective.
    }  
	\label{fig:main}
\end{center}
\vspace{-10pt}
\end{figure*}

\subsection{TimesBERT}

Transformers~\cite{vaswani2017attention} are currently the \emph{de facto} architecture of foundation models, especially for the vast use of GPT-style decoder-only Transformers in generative tasks. However, the primary objective of generative pre-trained models does not include learning bidirectional representations at different granularities. Thus, we adopt the BERT-style encoder-only architecture as a representation learning backbone. To cope with variable lengths in time points and variates, we design a unified time series embedding for structured representation learning.

\subsubsection{Time Series Embedding}

Given multivariate time series $\mathbf{X}=[\mathbf{x}_{1},\mathbf{x}_{2},...,\mathbf{x}_{C}]$ of $C$ variates, each variate $\mathbf{x}_{i}\in\mathbb{R}^{T}$ is a length-$T$ time series that will be divided into $N=\lceil \frac{T}{P} \rceil$ patches of patch length  $P$:
\begin{equation}
\begin{aligned}
    \mathbf{p}_{i,c}&=[\mathbf{X}_{(i-1)P+1,c},\dots,\mathbf{X}_{iP,c}],\\
\mathbf{z}^0_{i,c}&=\mathbf{W}_\texttt{in}\mathbf{p}_{i,c}+\mathbf{PE}_{i+Pc},
\end{aligned}
\end{equation}
where $\mathbf{W}_\texttt{in}\in\mathbb{R}^{D\times P}$ is a linear layer, and $ \mathbf{PE}_{i+Pc}\in\mathbb{R}^{D}$ denotes absolute position embedding. 
We adopt a shared learnable embedding $\mathbf{z}_{\texttt{[MASK]}}$ in each masked position. 

We repurpose the functional token $\texttt{[SEP]}$, which is used for next-sentence prediction in BERT, as a learnable embedding $\mathbf{z}_{\texttt{[VAR]}}$. 
Inspired by the functional token $\texttt{[CLS]}$ used for classifying a group of sentences in BERT, we also append a learnable embedding $\mathbf{z}_{\texttt{[DOM]}}$ at the beginning as the domain token. We formulate $\mathbf{Z}^0$ as the input of Transformer by stacking $(N+1)C+1$ token embeddings:
\begin{equation}\label{equ:structured}
\begin{aligned}
\mathbf{Z}^0&=\begin{pmatrix}
     \mathbf{z}_{\scalebox{0.5}{\texttt{[DOM]}}} & \mathbf{z}^0_{1:1} & \mathbf{z}^0_{2:1} & ... & \mathbf{z}_{\scalebox{0.5}{\texttt{[MASK]}}} & \mathbf{z}_{\scalebox{0.5}{\texttt{[VAR]}}} \\
     \cdot &\mathbf{z}^0_{1:2} & \mathbf{z}_{\scalebox{0.5}{\texttt{[MASK]}}} & \cdots & \mathbf{z}^0_{N,2} & \mathbf{z}_{\scalebox{0.5}{\texttt{[VAR]}}} \\
     \cdot &\vdots & \vdots & \ddots & \vdots & \vdots \\
     \cdot &\mathbf{z}^0_{1:C} & \mathbf{z}^0_{2:C} & \cdots & \mathbf{z}^0_{N,C} & \mathbf{z}_{\scalebox{0.5}{\texttt{[VAR]}}}
\end{pmatrix}.
\end{aligned}
\end{equation}

We implement packing~\cite{raffel2020exploring} to simultaneously train on different multivariate time series in one large context length (512 tokens in TimesBERT). In addition to aggregating global information, the functional token helps separate different training samples.

We adopt an encoder-only Transformer with dimensions $D$ and $L$ layers as the backbone of TimesBERT, which forwards the flattened token embeddings: 
\begin{equation}
\mathbf{Z}^l=\operatorname{TrmBlock}(\mathbf{Z}^{l-1}),\ l=1, \dots, L.
\end{equation}

As shown in Figure~\ref{fig:main}, we extend the 1D format of word sequences in BERT to accommodate multivariate time series with arbitrary variates and time points. In conjunction with the pre-training tasks designed, patterns at the patch level, variate level, and sample level are aggregated on the corresponding functional tokens, which ultimately forms multi-granularity representation extraction.

\subsection{Pre-training TimesBERT}

We design two pre-training objectives for structured time series to acquire a generic understanding. 

\subsubsection{Task \#1: Masked Patch Modeling}

Inspired by the masked language modeling task utilized in BERT, we employ masked patch modeling (MPM) to provide a pedestal understanding ability for the foundation model.
Given the input token sequence, we adopt a masked ratio $\alpha=25\%$ for non-functional tokens. To minimize the discrepancy between pre-training and fine-tuning tasks, these selected masked tokens are then actually replaced by $\mathbf{z}_{\texttt{[MASK]}}$ with a $90\%$ probability. Here let $\mathbf{z}_i$ denote reconstructed token at position $i$, let $S = \alpha NC$ denote the total number of masked patches, and we denote reconstructed patches as $\{\widehat{\mathbf{p}}_i\}_{i=1}^S$. We use a linear layer $\mathbf{W}_\texttt{out}\in\mathbb{R}^{D\times P}$ to project tokens to reconstructed patches:
\begin{equation}
\{\widehat{\mathbf{p}}_i\}_{i=1}^S = \{\mathbf{z}_i\}_{i=1}^S\mathbf{W}_\texttt{out}.
\end{equation}
Given the ground truth of masked patches as $\{\mathbf{p}_i\}_{i=1}^S$, the masked patch modeling objective is formulated as:
\begin{equation}\label{equ:objective_mpm}
\mathcal{L}_{\text{MPM}} = \frac{1}{SP}\sum\limits_{i=1}^{S}||\mathbf{p}_i-\widehat{\mathbf{p}}_i||_2^2.
\end{equation}
Equation~\ref{equ:objective_mpm} enhances the basic model capability to extract temporal representations from local variations. 
Nevertheless, ablation studies~\ref{subsect:model_analysis} indicate that MPM alone is insufficient to provide optimal transfer ability toward downstream tasks. For tasks requiring explicit understanding of global representations, such as classification and anomaly detection, it is necessary to propose a pre-training task that better aligns with the document-like structure of time series data.

\subsubsection{Task \#2: Functional Token Prediction}

Despite the goal of masked patch modeling to model temporal patterns in a single time series, it struggles to explicitly handle inter-variate relationships and effectively aggregate the overall characteristics of variates. Inspired by next sentence prediction (NSP) task in BERT, we propose \emph{Functional Token Prediction} (FTP) relying on special tokens. 

We design a variate discrimination task. Given a multivariate time series with \(C \geq 2\), we randomly replace one variate with another from another dataset. The task of the model is to identify the replaced variate by its own variation patterns. Here let $\mathbf{z}_{\mathbf{v}_c}$ denote the output of $\mathbf{z}_{\texttt{[VAR]}}$ of $n$-th variate, and we project $\mathbf{z}_{\mathbf{v}_c}$ with a linear layer \(\mathbf{W}_\texttt{VAR} \in \mathbb{R}^{D \times 2}\) to classify whether a variate originated the same as other variates.

Here let $\mathbf{z}_{g}$ denote the output of $\mathbf{z}_{\texttt{[DOM]}}$. Based on the domain token \(\mathbf{z}_g\), we propose a domain classification task. With \(M\) datasets indexed in pre-training, the backbone provides outputs, and \(\mathbf{z}_g\) is fed into a linear layer \(\mathbf{W}_\texttt{DOM} \in \mathbb{R}^{D \times M}\) to predict the dataset index of the series.

Based on the aforementioned process, the functional token prediction objective can be formulated as follows: 
\begin{equation}\label{equ:objective_ftp}
\mathcal{L}_{\scalebox{0.5}{\text{FTP}}} = - \sum_{c=1}^C \sum_{i=1}^2 \mathbf{y}^c_i\log (\mathbf{z}_{\mathbf{v}_c}\mathbf{W}_{\scalebox{0.3}{\texttt{VAR}}})_i -  \sum_{i=1}^M \mathbf{y}^d_i\log (\mathbf{z}_g\mathbf{W}_{\scalebox{0.3}{\texttt{DOM}}})_i,
\end{equation}
where the one-hot vector $\mathbf{y}^c$ denotes labels marking whether $\mathbf{v}_c$ is the replaced variate, and the one-hot vector $\mathbf{y}^d$ denotes the index of the dataset. 

Finally, the training objective is represented as follows:
\begin{equation}\label{equ:objective}
\mathcal{L} = \mathcal{L}_{\text{MPM}} + \mathcal{L}_{\text{FTP}}.
\end{equation}
Our functional token prediction task treats each variate as a time series sentence, requiring them to distribute and aggregate with one another to identify their similarities and differences in relation to the entire sequence. As functional tokens learn representations at varying granularities, they enhance task versatility during downstream adaptation, allowing a series of token embeddings to be employed for understanding time series patches, variables, and domains. Following the pre-training phase, the task head is removed, while the Transformer backbone is adapted for representation extraction during fine-tuning. This process effectively decouples the pre-trained backbone from the task design.

\subsection{Pre-Training Data}

We construct large-scale time series corpora from various sources. We adopt the LOTSA dataset~\cite{woo2024unified} as the main body of the pre-training dataset, taking into account the needs of the basic model for multi-domain and pattern diversity requirements. Simultaneously, there is a notable discrepancy in data features between understanding-oriented domains, such as medical~\cite{gow2023mimic}, and pre-training data used for forecasting, thus to account for the varying temporal dynamics and variate correlations of time series across different tasks, we incorporate the UEA Archive~\cite{Bagnall2018TheUM} to achieve a balanced data portrait, forming a large-scale corpus with a total of 260 billion time points. Based on our structure-preserving design for multivariate time series, TimesBERT fully leverages time series native during large-scale pre-training to achieve rapid and effective transfer for complex time series tasks.

\subsection{Fine-Tuning TimesBERT}

Analogous to the fine-tuning methodology employed with BERT, we adopt a trainable output layer during the fine-tuning phase of TimesBERT to accommodate various downstream datasets. Considering the low er information density, we utilize all tokens to ensure a comprehensive representation when migrating to the classification, while tokens at the corresponding positions are directly used as representations for imputation and anomaly detection. For diverse understanding tasks, the fine-tuning paradigm of BERT-style models empirically exhibits advantages compared to the prevailing zero-shot paradigm of GPT-style models.

\section{Experiments}

In order to validate the capacity of TimesBERT for typical understanding scenarios, we conduct experiments in time series classification, imputation, short-term forecasting, and anomaly detection tasks.  We compare TimesBERT with state-of-the-art task-specific and general models and exhibit the benefit of pre-training.
In addition,  we conducted comprehensive ablation studies to evaluate various aspects of the model design and its capabilities.
We provide implementation details and model configurations in Appendix~\ref{sec:implementation_detail}.

\begin{figure}[htbp]
\begin{center}
    \centerline{\includegraphics[width=1\columnwidth]{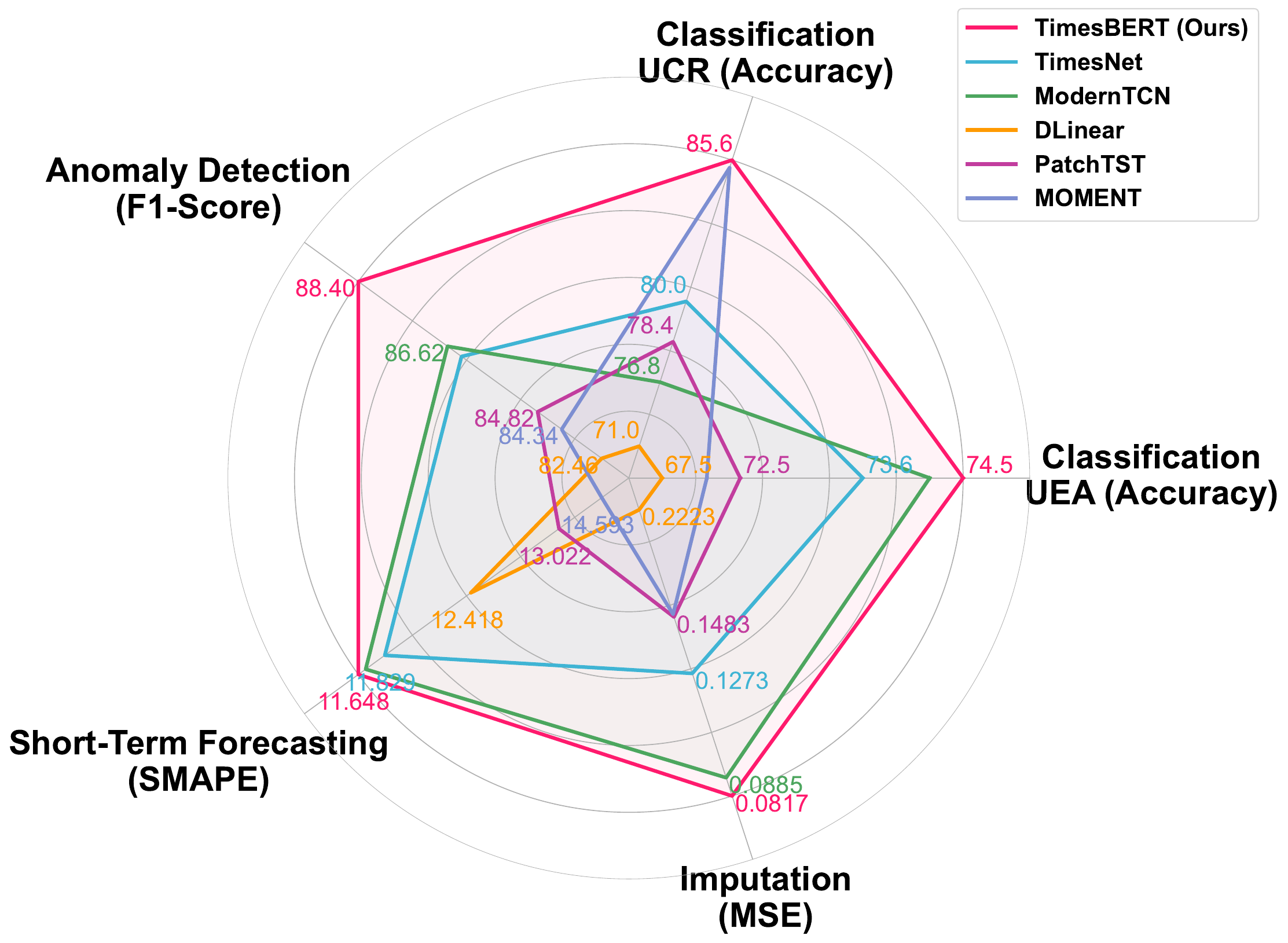}}
    \vspace{-10pt}
	\caption{Overall Performance of TimesBERT.} 
	\label{fig:radar}
\end{center}
\vspace{-20pt}
\end{figure}

\subsection{Classification}

\paragraph{Setups} Time series classification represents a typical data understanding task. During the feature extraction process of the pre-trained model, capturing temporal patterns necessitates that the classifier possesses robust global comprehension capabilities. We utilize two benchmark datasets for time series classification. Specifically, we employ 10 subsets from the UEA Archive~\cite{Bagnall2018TheUM} and 91 subsets from the UCR Archive~\cite{dau2019ucr}, spanning diverse domains such as biology, physics, environmental monitoring, human activity recognition, and finance, among others. These datasets encompass varying variate numbers and sequence lengths which exhibit diversity in granularity.

\begin{figure*}[ht]
\begin{center}
    \center{\includegraphics[width=\textwidth]{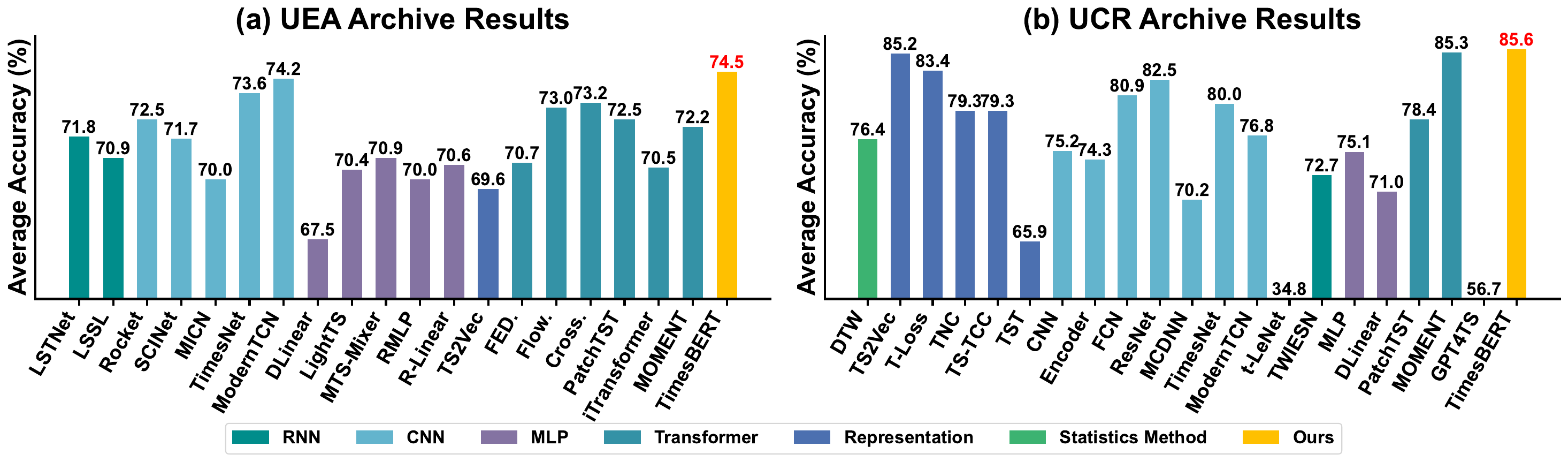}}
    \vspace{-18pt}
	\caption{Averaged results of classification task on UEA and UCR Archive. We illustrate an extensive range of analytical baselines, among which TimesBERT achieved state-of-the-art performance in two classification benchmarks. The results of previous methods are directly taken down from previous literature for fair comparison. See Table \ref{tab:classification_uea_full} and Table \ref{tab:classification_ucr_full} for full results.} 
	\label{fig:classification}
\end{center}
\vspace{-5pt}
\end{figure*}

\paragraph{Results} Figure~\ref{fig:classification} illustrates the inference outcomes of our model on UEA and UCR Archive. Compared to statistical methods, state-of-the-art deep learning models, and classical unsupervised representation learning methods, the average classification accuracy of TimesBERT shows consistent performance enhancements. Notably, given the substantial variations in sequence length, number of variates, change patterns, and class counts across existing time series classification benchmarks, TimesBERT exhibits comprehensive improvements across over one hundred benchmark datasets.

\subsection{Anomaly Detection}

\paragraph{Setups}
Time series anomaly detection is a widely discussed task aimed at discerning anomalous data segments, which is of great importance in actual time series analysis. We include five widely-used anomaly detection benchmarks into our experiments, namely: SMD~\cite{Su2019RobustAD}, MSL~\cite{Hundman2018DetectingSA}, SMAP~\cite{Hundman2018DetectingSA}, SWaT~\cite{DBLP:conf/cpsweek/MathurT16}, PSM~\cite{DBLP:conf/kdd/AbdulaalLL21}. Our evaluation follows the unsupervised time series anomaly detection logic mentioned in previous works such as TimesNet~\cite{wu2022timesnet}, where datasets are split into non-overlapping sliding windows, and the reconstruction error is applied as the anomaly criterion. 

\paragraph{Results}

Our experiments result in table~\ref{tab:timesbert_anomaly_detection} shows that TimesBERT performs better than previous sota baselines such as TimesNet~\cite{wu2022timesnet}. We highlight that TimesBERT's improvements are consistent across all time series anomaly detection benchmarks, which demonstrates the robust adaptability of pre-trained models to complex downstream understanding tasks and diverse datasets.

\begin{table*}[htbp]
  \caption{Anomaly detection results. We calculate the F1-score (as \%) for each dataset. *. means the *former. A higher value of the F1-score indicates a better performance. See Table \ref{tab:full_anomaly_results} for full results.}\label{tab:anomaly_results}
  \vskip 0.05in
  \label{tab:timesbert_anomaly_detection}
  \centering
  \begin{small}
  \begin{sc}
  \renewcommand{\multirowsetup}{\centering}
  \setlength{\tabcolsep}{2.8pt}
  \resizebox{\linewidth}{!}{
  \begin{tabular}{l|ccccccccccccccc}
    \toprule
    \multicolumn{1}{c}{\multirow{2}{*}{Model}} & 
    \multicolumn{1}{c}{\rotatebox{0}{\scalebox{0.75}{\textbf{TimesBERT}}}} &
    \multicolumn{1}{c}{\rotatebox{0}{\scalebox{0.75}{MTCN}}} &   
    \multicolumn{1}{c}{\rotatebox{0}{\scalebox{0.75}{TimesNet}}} &
    \multicolumn{1}{c}{\rotatebox{0}{\scalebox{0.75}{ETS.}}} &
    \multicolumn{1}{c}{\rotatebox{0}{\scalebox{0.75}{FED.}}} &
    \multicolumn{1}{c}{\rotatebox{0}{\scalebox{0.75}{LightTS}}} &
    \multicolumn{1}{c}{\rotatebox{0}{\scalebox{0.75}{DLinear}}} &
     \multicolumn{1}{c}{\rotatebox{0}{\scalebox{0.75}{NS.}}} &
     \multicolumn{1}{c}{\rotatebox{0}{\scalebox{0.75}{Auto.}}} &
     \multicolumn{1}{c}{\rotatebox{0}{\scalebox{0.75}{Pyra.}}} &  
     \multicolumn{1}{c}{\rotatebox{0}{\scalebox{0.75}{Anomaly.}}} & 
     \multicolumn{1}{c}{\rotatebox{0}{\scalebox{0.75}{Informer}}} &
     \multicolumn{1}{c}{\rotatebox{0}{\scalebox{0.75}{Reformer}}} &
     \multicolumn{1}{c}{\rotatebox{0}{\scalebox{0.75}{LogTrans}}} & \multicolumn{1}{c}{\rotatebox{0}{\scalebox{0.75}{Trans.}}}    \\
    \multicolumn{1}{c}{} & \multicolumn{1}{c}{\scalebox{0.7}{(\textbf{Ours})}} &
    \multicolumn{1}{c}{\scalebox{0.75}{\citeyearpar{donghao2024moderntcn}}} &
    \multicolumn{1}{c}{\scalebox{0.75}{\citeyearpar{wu2022timesnet}}} &
    \multicolumn{1}{c}{\scalebox{0.75}{\citeyearpar{woo2022etsformer}}} &
    \multicolumn{1}{c}{\scalebox{0.75}{\citeyearpar{zhou2022fedformer}}} &
    \multicolumn{1}{c}{\scalebox{0.75}{\citeyearpar{Zhang2022LessIM}}} &
    \multicolumn{1}{c}{\scalebox{0.75}{\citeyearpar{Zeng2022AreTE}}} & \multicolumn{1}{c}{\scalebox{0.75}{\citeyearpar{Liu2022NonstationaryTR}}} & \multicolumn{1}{c}{\scalebox{0.75}{\citeyearpar{wu2021autoformer}}} & \multicolumn{1}{c}{\scalebox{0.75}{\citeyearpar{liu2021pyraformer}}} &  \multicolumn{1}{c}{\scalebox{0.75}{\citeyearpar{xu2021anomaly}}} &\multicolumn{1}{c}{\scalebox{0.75}{\citeyearpar{haoyietal-informer-2021}}} & \multicolumn{1}{c}{\scalebox{0.75}{\citeyearpar{kitaev2020reformer}}}& \multicolumn{1}{c}{\scalebox{0.75}{\citeyearpar{2019Enhancing}}} & \multicolumn{1}{c}{\scalebox{0.75}{\citeyearpar{NIPS2017_3f5ee243}}}   \\
    \toprule
    \scalebox{0.9}{SMD} & \boldres{\scalebox{0.9}{86.04}} & \secondres{\scalebox{0.9}{85.81}} & \secondres{\scalebox{0.9}{85.81}} & \scalebox{0.9}{83.13} & \scalebox{0.9}{85.08} & \scalebox{0.9}{82.53} & \scalebox{0.9}{77.10} & \scalebox{0.9}{84.72} & \scalebox{0.9}{85.11} & \scalebox{0.9}{83.04} & \scalebox{0.9}{85.49} & \scalebox{0.9}{81.65} & \scalebox{0.9}{75.32} & \scalebox{0.9}{76.21} & \scalebox{0.9}{79.56}  \\
    \scalebox{0.9}{MSL} & \boldres{\scalebox{0.9}{88.07}} & \scalebox{0.9}{84.92} & \secondres{\scalebox{0.9}{85.15}} & \scalebox{0.9}{85.03} & \scalebox{0.9}{78.57} & \scalebox{0.9}{78.95} & {\scalebox{0.9}{84.88}} & \scalebox{0.9}{77.50} & \scalebox{0.9}{79.05} & \scalebox{0.9}{84.86} & \scalebox{0.9}{83.31} & \scalebox{0.9}{84.06} & \scalebox{0.9}{84.40} & \scalebox{0.9}{79.57} & \scalebox{0.9}{78.68} \\
    \scalebox{0.9}{SMAP} & \boldres{\scalebox{0.9}{75.69}} & \scalebox{0.9}{71.26} & \secondres{\scalebox{0.9}{71.52}}& \scalebox{0.9}{69.50} & \scalebox{0.9}{70.76} & \scalebox{0.9}{69.21} & \scalebox{0.9}{69.26} & \scalebox{0.9}{71.09} & \scalebox{0.9}{71.12} & \scalebox{0.9}{71.09} & \scalebox{0.9}{71.18} & \scalebox{0.9}{69.92} & \scalebox{0.9}{70.40} & \scalebox{0.9}{69.97} & \scalebox{0.9}{69.70} \\
    \scalebox{0.9}{SWaT} & \boldres{\scalebox{0.9}{93.95}} & \secondres{\scalebox{0.9}{93.86}} & \scalebox{0.9}{91.74}& \scalebox{0.9}{84.91} & \scalebox{0.9}{93.19} & \scalebox{0.9}{93.33} & \scalebox{0.9}{87.52} & \scalebox{0.9}{79.88} & \scalebox{0.9}{92.74} & \scalebox{0.9}{91.78} & \scalebox{0.9}{83.10} & \scalebox{0.9}{81.43} & \scalebox{0.9}{82.80} & \scalebox{0.9}{80.52} & \scalebox{0.9}{80.37} \\
    \scalebox{0.9}{PSM} & \boldres{\scalebox{0.9}{98.27}} & \scalebox{0.9}{97.23} & \secondres{\scalebox{0.9}{97.47}}& \scalebox{0.9}{91.76} & \scalebox{0.9}{97.23} & \scalebox{0.9}{97.15} & \scalebox{0.9}{93.55} & \scalebox{0.9}{97.29} & \scalebox{0.9}{93.29} & \scalebox{0.9}{82.08} & \scalebox{0.9}{79.40} & \scalebox{0.9}{77.10} & \scalebox{0.9}{73.61} & \scalebox{0.9}{76.74} & \scalebox{0.9}{76.07} \\
    \midrule
    \scalebox{0.9}{Avg. F1} & \boldres{\scalebox{0.9}{88.40}} & \secondres{\scalebox{0.9}{86.62}} & \scalebox{0.9}{86.34}& \scalebox{0.9}{82.87} & \scalebox{0.9}{84.97} & \scalebox{0.9}{84.23} & \scalebox{0.9}{82.46} & \scalebox{0.9}{82.08} & \scalebox{0.9}{84.26} & \scalebox{0.9}{82.57} & \scalebox{0.9}{80.50} & \scalebox{0.9}{78.83} & \scalebox{0.9}{77.31} & \scalebox{0.9}{76.60} & \scalebox{0.9}{76.88} \\
    \bottomrule
  \end{tabular}
  }
  \vspace{-5pt}
  \end{sc}
  \end{small}
\end{table*}

\subsection{Imputation}

\paragraph{Setups} Given the pervasive occurrence of missing values in real-world industrial production scenarios, we evaluate the effectiveness of time series imputation tasks, where bidirectional information is significantly important for enhancing the model's ability to analyze missing segments. Additionally, this task necessitates that the model comprehends and encapsulates the overall features of the series through high-order representations, revealing the advantages of TimesBERT. Since value missing in real scenarios often occurs in continuous segments, we employ patch-level imputation for evaluation following Timer~\cite{liu2024timer}, which is more challenging than point-level imputation.

\begin{figure}[t]
\begin{center}
    \centerline{\includegraphics[width=1\columnwidth]{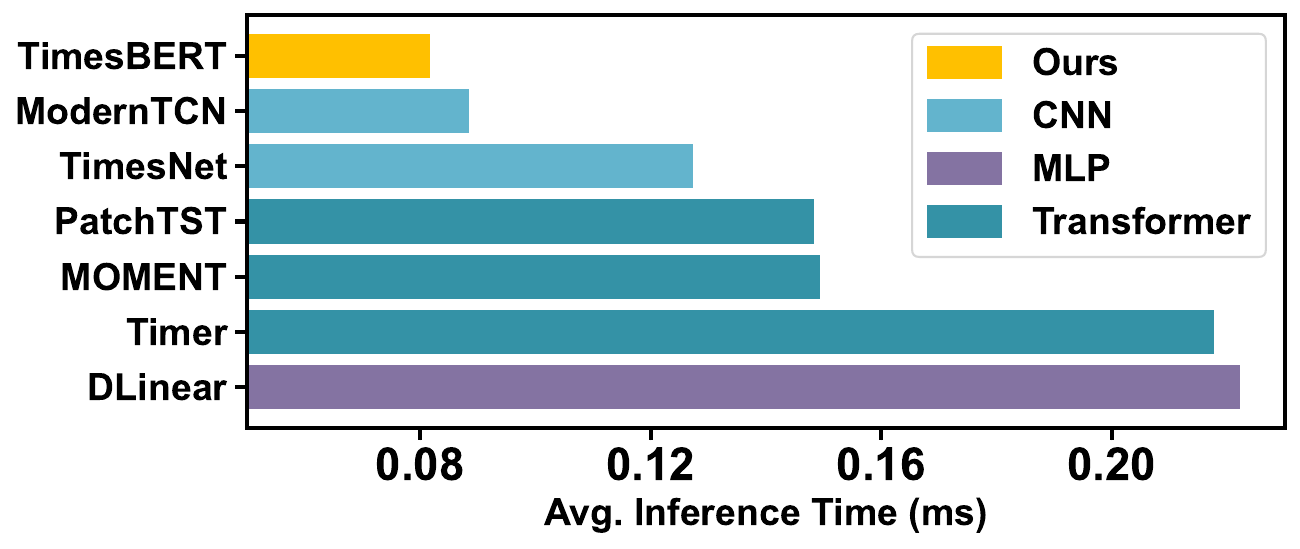}}
    \vspace{-10pt}
	\caption{Averaged results of imputation tasks. We randomly mask $\{12.5\%, 25\%, 37.5\%, 50\%\}$ patches. The results are averaged from 4  mask ratios. See Table  \ref{tab:imputation_p100} and Table \ref{tab:imputation_timesbert} for full results.} 
	\label{fig:imputation_baseline}
\end{center}
\vspace{-20pt}
\end{figure}

\begin{figure}[t]
\begin{center}
    \centerline{\includegraphics[width=1\columnwidth]{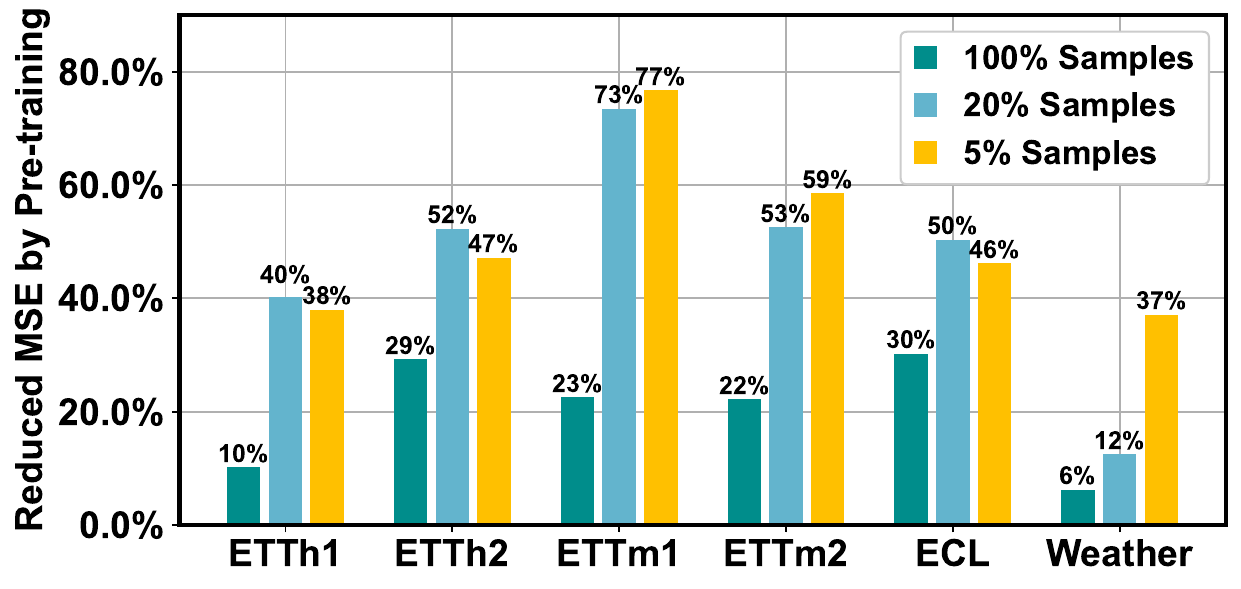}}
    \vspace{-10pt}
	\caption{Pre-training benefit of TimesBERT on the downstream imputation task with $100\%$, $20\%$ and $5\%$ available samples. Each dataset is imputed with four mask ratios and we calculate the average reduced imputation error in MSE relative to training from scratch. See Table \ref{tab:imputation_p100}, Table \ref{tab:imputation_p20} and Table \ref{tab:imputation_p5} for full results.} 
	\label{fig:imputation}
\end{center}
\vspace{-20pt}
\end{figure}

\paragraph{Results}  We conduct comprehensive evaluations on selected six classical benchmarks with four different mask ratios in order to avoid data leakage caused by pre-trained corpus and compared them with advanced general and foundation models. As shown in Table~\ref{fig:imputation_baseline}, TimesBERT achieves a $7.7\%$ loss reduction compared to the state-of-the-art model on this task. In addition, we test the benefits of model pre-training under the data scarcities of $\{5\%, 20\%, 100\%\}$. As shown in Figure~\ref{fig:imputation}, the pre-trained model gives a significant improvement with fewer fine-tuning samples.

\subsection{Short-Term Forecasting}

\paragraph{Setups} Short-term time series forecasting is extensively utilized in domains such as meteorological prediction, market analysis, and finance. Unlike long-term forecasting tasks, which depend on capturing consistent local change patterns within the retrospective window and require robust model roll-out capabilities, short-term forecasting emphasizes providing trend predictions for the forecast horizon based on the overall characteristics of the series. Consequently, it is more suitable for models with understanding capabilities. For this task, we employ the M4 dataset~\cite{M4team2018dataset} as a benchmark. We adhere to the evaluation established by TimesNet~\cite{wu2022timesnet}.

\begin{table*}[htbp]
  \caption{Average short-term forecasting results on the M4~\cite{M4team2018dataset}. Full results are provided in Table \ref{tab:short_term_forecast_full}.}
  \label{tab:short_term_forecast}
  \vspace{-5pt}
  \vskip 0.15in
  \centering
  \begin{small}
  \begin{sc}
  \renewcommand{\multirowsetup}{\centering}
  \setlength{\tabcolsep}{2pt}
  \resizebox{\linewidth}{!}{
  \begin{tabular}{l|c|cccccccccccccccccc}
    \toprule
    \multicolumn{2}{l|}{\multirow{2}{*}{Method}} & \scalebox{0.75}{\textbf{TimesBERT}} & \scalebox{0.75}{MTCN} & \scalebox{0.75}{TimesNet} & \scalebox{0.75}{iTrans.} & \scalebox{0.75}{Koopa} & \scalebox{0.75}{NHiTS} & \scalebox{0.75}{DLinear} & \scalebox{0.75}{PatchTST} & \scalebox{0.75}{MICN} & \scalebox{0.75}{TiDE} & \scalebox{0.75}{MOMENT} & \scalebox{0.75}{NBEATS}  \\ 
    \multicolumn{2}{l|}{} & \scalebox{0.75}{\textbf{(Ours)}} & \scalebox{0.75}{\citeyearpar{wu2022timesnet}} & \scalebox{0.75}{\citeyearpar{donghao2024moderntcn}} & \scalebox{0.75}{\citeyearpar{liu2023itransformer}} & \scalebox{0.75}{\citeyearpar{liu2023koopa}} & \scalebox{0.75}{\citeyearpar{challu2023nhits}} & \scalebox{0.75}{\citeyearpar{zeng2023transformers}}  & \scalebox{0.75}{\citeyearpar{nie2022time}} &  \scalebox{0.75}{\citeyearpar{wang2022micn}} & \scalebox{0.75}{\citeyearpar{das2023long}} & \scalebox{0.75}{\citeyearpar{goswami2024moment}} & \scalebox{0.75}{\citeyearpar{oreshkin2019n}} \\
    \toprule
    \multirow{3}{*}{\scalebox{0.8}{\rotatebox{90}{Average}}}
     & \scalebox{0.8}{SMAPE} & \boldres{\scalebox{0.8}{11.648}} & \secondres{\scalebox{0.8}{11.698}} & \scalebox{0.8}{11.829} & \scalebox{0.8}{12.684} & \scalebox{0.8}{11.863} & \scalebox{0.8}{11.960} & \scalebox{0.8}{12.418} & \scalebox{0.8}{13.022} & \scalebox{0.8}{13.023} & \scalebox{0.8}{13.950} & \scalebox{0.8}{14.593} & \scalebox{0.8}{11.910} \\
     & \scalebox{0.8}{MASE}  & \secondres{\scalebox{0.8}{1.560}} & \boldres{\scalebox{0.8}{1.556}} & \scalebox{0.8}{1.585}  & \scalebox{0.8}{1.764}  & \scalebox{0.8}{1.595}  & \scalebox{0.8}{1.606}  & \scalebox{0.8}{1.656}  & \scalebox{0.8}{1.814}  & \scalebox{0.8}{1.836}  & \scalebox{0.8}{1.940}  & \scalebox{0.8}{2.161}  & \scalebox{0.8}{1.613}  \\
     & \scalebox{0.8}{OWA}   & \boldres{\scalebox{0.8}{0.837}} & \secondres{\scalebox{0.8}{0.838}} & \scalebox{0.8}{0.851} & \scalebox{0.8}{0.929}  & \scalebox{0.8}{0.858}  & \scalebox{0.8}{0.861}  & \scalebox{0.8}{0.891}  & \scalebox{0.8}{0.954}  & \scalebox{0.8}{0.960}  & \scalebox{0.8}{1.020}  & \scalebox{0.8}{1.103}  & \scalebox{0.8}{0.862} \\
    \bottomrule
  \end{tabular}
  }
    \vspace{-21pt}
  \end{sc}
  \end{small}
\end{table*}

\paragraph{Results}

We evaluate well-acknowledged forecasting models, including iTransformer~\cite{liu2023itransformer}, PatchTST~\cite{nie2022time} and TimesNet~\cite{wu2022timesnet}, by three widely accepted metrics on the M4 dataset. As shown in Table~\ref{tab:short_term_forecast}, TimesBERT outperforms previous models consistently on all average SMAPE and OWA.

\subsection{Model Analysis}\label{subsect:model_analysis}

\paragraph{Pre-training Tasks} During the pre-training phase, we employ two distinct pre-training tasks and concurrently optimize three task-specific heads, among which there is explicit complementary relation. We conduct an ablation study on the tasks and compare in detail the performance of pre-trained TimesBERT with and without the functional token prediction task before transferring to time series classification tasks. On this basis, we examine the impact of functional token selection during the fine-tuning stage.

As shown in Table~\ref{tab:pre_train}, on average, all our pre-training tasks and functional tokens while fine-tuning yield positive improvements. Notably, the inclusion of functional tokens provides a significant boost to classification performance, confirming their ability to aggregate relevant representation.

\begin{table}[htbp]
\caption{Ablation of pre-training tasks. D in the ``pre-train'' row means using domain classification, and V means using variate discrimination; In the ``fine-tuning'' row, D means $\mathbf{z}_{\texttt{[DOM]}}$ is used for the task head while fine-tuning, and V means $\mathbf{z}_{\texttt{[VAR]}}$ is used.}
\label{tab:pre_train}
\setlength{\tabcolsep}{6pt}
\vskip -0.8in
\begin{center}
\begin{small}
\begin{sc}
\begin{tabular}{l|ccc|c|c}
\toprule
Pre-train & \multicolumn{3}{c|}{D+V}     & D       & None    \\
\midrule
Fine-tune & D+V     & D       & None    & D       & None    \\
\midrule
EC        & \boldres{34.60} & \secondres{34.22} & 33.46 & 30.04 & 32.32 \\
FD        & 68.67 & 68.67 & \secondres{69.44} & 68.93 & \boldres{69.55} \\
HW        & \boldres{36.59} & \boldres{36.59} & \secondres{36.12} & 35.76& 35.18 \\
HB        & \boldres{78.54} & \boldres{78.54} & 77.56 & \secondres{78.05} & \boldres{78.54} \\
JV        & \secondres{97.57} & \secondres{97.57} & \boldres{98.38} & \boldres{98.38} & \boldres{98.38} \\
SRS1      & \secondres{93.17} & \secondres{93.17} & \boldres{93.52} & 91.47 & 90.78 \\
SRS2      & \secondres{58.33} & \secondres{58.33} & 57.22 & \boldres{59.44} & 57.22 \\
SWJ       & \secondres{99.41} & \secondres{99.41} & \boldres{99.50} & 99.27 & 99.32 \\
SW        & \secondres{95.00} & \secondres{95.00} & \boldres{95.31} & 94.38 & 93.44 \\
\midrule
Avg.       & \boldres{73.54} & \secondres{73.50} & 73.39 & 72.86 & 72.75\\
\bottomrule
\end{tabular}
\vspace{-5pt}
\end{sc}
\end{small}
\end{center}
\vskip -0.2in
\end{table}

\paragraph{Initialization} Recent studies on large language models and vision models for time series~\cite{ zhou2023one, liu2024autotimes, chen2024visionts} have demonstrated the advantages of leveraging pre-trained models on other modalities for time series modeling to some extent, while we posit that time series exhibit more complex intrinsic variation patterns. Therefore, in this ablation study, we attempt to directly initialize TimesBERT using a pre-trained BERT model and compare its performance with that of a normal pre-trained TimesBERT on classification tasks. 

\begin{figure}[t]
\begin{center}
    \centerline{\includegraphics[width=1\columnwidth]{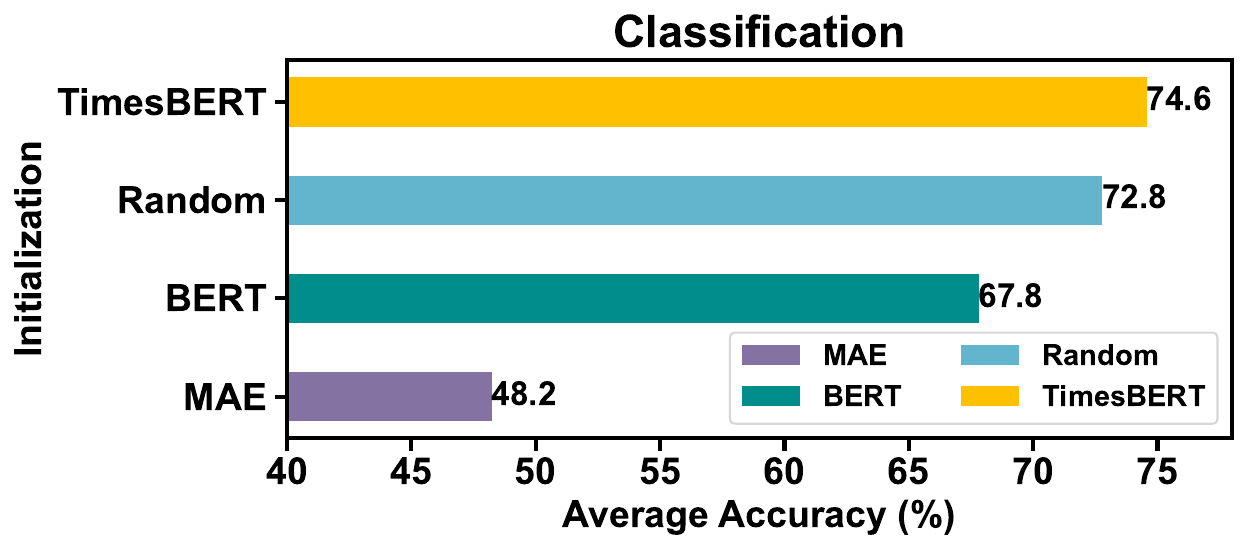}}
    \vspace{-10pt}
	\caption{We initialize TimesBERT for downstream tasks using three methods: (1) pre-trained TimesBERT model, (2) random initialization, (3) A Vision Transformer~\cite{dosovitskiy2020image} pre-trained with Masked Autoencoder ~\cite{he2022masked}, and (4) a BERT~\cite{devlin2018bert} model pre-trained on language.} 
	\label{fig:initialization}
\end{center}
\vspace{-25pt}
\end{figure}

Figure~\ref{fig:initialization} illustrates the inclusion of different initialization methods alongside a random initialization as a control group, which highlights the fundamental difference among linguistic manifolds, image semantic space, and time series, while also demonstrating the significant improvements achieved by our pre-trained models. These findings underscore the importance of native time-series pre-training.

\paragraph{Multivariate Modeling} Time series understanding tasks typically require robust multivariate modeling capabilities, in which capturing multivariate correlations is often essential for accurately representing the overall data features. Nevertheless, existing foundation model designs often employ Channel Independence (CI) to avoid interference from variate modeling~\cite{liu2024timer,ansari2024chronos}. We conduct ablation studies on classification and imputation tasks utilizing datasets with explicit multivariate features.

\begin{figure}[t]
\begin{center}
    \centerline{\includegraphics[width=1\columnwidth]{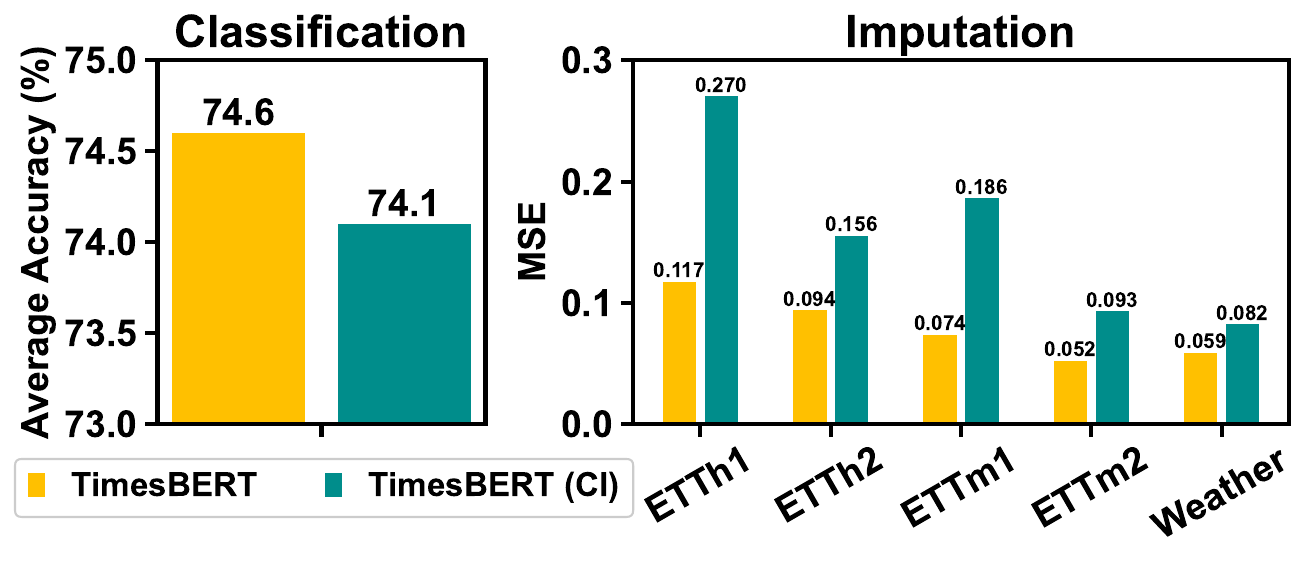}}
    \vspace{-10pt}
	\caption{Ablation of TimesBERT's multivariate capabilities on classification and imputation tasks. CI indicates that the model uses the Channel Independence strategy~\cite{nie2022time}.} 
	\label{fig:multivariate}
\end{center}
\vspace{-30pt}
\end{figure}

Figure~\ref{fig:multivariate} indicates that in both tasks, compared to employing CI to mitigate the interference from variate relationships, TimesBERT consistently leverages the variate correlations to achieve benefits while supporting multivariate inputs.

\section{Conclusion and Future Work}

In this paper, we highlight that multivariate time series and multisentence text documents exhibit a similar multi-granularity structure. Inspired by BERT which facilitates structured representation learning for agnostic downstream tasks, we leverage BERT-style architectures for generic time series understanding, which is achieved by repurposing masked modeling and functional token prediction for arbitrary multivariate time series. By large-scale pre-training on 260 billion time points across different domains, TimesBERT surpasses state-of-the-art models across four typical understanding tasks, which validates the exceptional generalization capabilities that BERT can also offer in the realm of time series analysis. We will explore the adaption of functional tokens and delve into domain-universal pre-training for more time series understanding tasks.

\bibliography{example_paper}
\bibliographystyle{icml2025}

\newpage
\appendix
\onecolumn

\section{Implementation Details}\label{sec:implementation_detail}
\subsection{Pre-training}
All experiments in this paper are implemented in PyTorch~\cite{Paszke2019PyTorchAI} and conducted on NVIDIA 4090 GPU. We used AdamW as the optimizer in the pre-training phase with $\beta_1=0.9, \beta_2=0.99$, and applied the cosine annealing algorithm for learning rate decay. Specifically, we use $1\times 10^{-4}$ as the initial learning rate and $2\times 10^{-7}$ as the final learning rate. To conduct fair ablation experiments, all our pre-trained models are trained for $30000$ steps on a large-scale time series corpus, where the batch size is $40\times 8=320$ and the context length of each sequence, that is, the number of patches, is $512$. To improve the convergence speed, we use the packing~\cite{raffel2020exploring} for large-scale pre-training.

For alignment with BERT~\cite{devlin2018bert}, our pre-trained model size is the same as $\mathbf{BERT}_{\mathbf{base}}$, i.e., we denote the number of layers as $L$, the hidden size as $H$, and the number of self-attention heads as $A$, then the model size is $L=12, H=768, A=12$, and the total number of parameters is $85.6$M. Considering the diversity of downstream datasets, we use models with different patch lengths on different tasks to adapt to diverse understanding tasks, including $36$ for classification, $24$ for imputation, and $4$ for short-term forecasting and anomaly detection tasks.

\subsection{Downstream Tasks}

\begin{table}[thbp]
  \vspace{-10pt}
  \caption{Dataset descriptions. The dataset size is organized in (Train, Validation, Test). * means that different subsets have different values}\label{tab:downstream_dataset}
  \vskip 0.1in
  \centering
  \begin{threeparttable}
  \begin{sc}
  \begin{scriptsize}
  \renewcommand{\multirowsetup}{\centering}
  \setlength{\tabcolsep}{12pt}
  \begin{tabular}{l|l|c|c|c|c}
    \toprule
    Tasks & Dataset & Variate & Series Length & Dataset Size & \scalebox{0.8}{Information (Frequency)} \\
    \toprule
     & M4-Yearly & 1 & 6 & (23000, 0, 23000) & \scalebox{0.8}{Demographic} \\
    \cmidrule{2-5}
     & M4-Quarterly & 1 & 8 & (24000, 0, 24000) & \scalebox{0.8}{Finance} \\
    \cmidrule{2-5}
    Forecasting & M4-Monthly & 1 & 18 & (48000, 0, 48000) & \scalebox{0.8}{Industry} \\
    \cmidrule{2-5}
    (short-term) & M4-Weakly & 1 & 13 & (359, 0, 359) & \scalebox{0.8}{Macro} \\
    \cmidrule{2-5}
     & M4-Daily & 1 & 14 & (4227, 0, 4227) & \scalebox{0.8}{Micro} \\
    \cmidrule{2-5}
     & M4-Hourly & 1 &48 & (414, 0, 414) & \scalebox{0.8}{Other} \\
    \midrule
    \multirow{5}{*}{Imputation} & ETTm1, ETTm2 & 7 & 192 & (34465, 11521, 11521) & \scalebox{0.8}{Electricity (15 mins)} \\
    \cmidrule{2-6}
     & ETTh1, ETTh2 & 7 & 192 & (8545, 2881, 2881) & \scalebox{0.8}{Electricity (15 mins)}\\
    \cmidrule{2-6}
     & Electricity & 321 & 192 & (18317, 2633, 5261) & \scalebox{0.8}{Electricity (15 mins)}\\
    \cmidrule{2-6}
    & Weather & 21 & 192 & (36792, 5271, 10540) & \scalebox{0.8}{Weather (10 mins)} \\
    \midrule
     & \scalebox{0.8}{EthanolConcentration} & 3 & 1751 & (261, 0, 263) & \scalebox{0.8}{Alcohol Industry}\\
    \cmidrule{2-6}
    & \scalebox{0.8}{FaceDetection} & 144 & 62 & (5890, 0, 3524) & \scalebox{0.8}{Face (250Hz)}\\
    \cmidrule{2-6}
    & \scalebox{0.8}{Handwriting} & 3 & 152 & (150, 0, 850) & \scalebox{0.8}{Handwriting}\\
    \cmidrule{2-6}
    & \scalebox{0.8}{Heartbeat} & 61 & 405 & (204, 0, 205)& \scalebox{0.8}{Heart Beat}\\
    \cmidrule{2-6}
    Classification & \scalebox{0.8}{JapaneseVowels} & 12 & 29 & (270, 0, 370) & \scalebox{0.8}{Voice}\\
    \cmidrule{2-6}
     & \scalebox{0.8}{PEMS-SF} & 963 & 144 & (267, 0, 173) & \scalebox{0.8}{Transportation (Daily)}\\
    \cmidrule{2-6}
    & \scalebox{0.8}{SelfRegulationSCP1} & 6 & 896 & (268, 0, 293) & \scalebox{0.8}{Health (256Hz)}\\
    \cmidrule{2-6}
    & \scalebox{0.8}{SelfRegulationSCP2} & 7 & 1152 & (200, 0, 180) & \scalebox{0.8}{Health (256Hz)}\\
    \cmidrule{2-6}
    & \scalebox{0.8}{SpokenArabicDigits} & 13 & 93 & (6599, 0, 2199) & \scalebox{0.8}{Voice (11025Hz)}\\
    \cmidrule{2-6}
    & \scalebox{0.8}{UWaveGestureLibrary} & 3 & 315 & (120, 0, 320) & \scalebox{0.8}{Gesture}\\
    \cmidrule{2-6}
    & \scalebox{0.8}{UCR Archive} & 1 & * & (*, 0, *) & *\\
    \midrule
     & SMD & 38 & 40 & (566724, 141681, 708420) & \scalebox{0.8}{Server Machine} \\
    \cmidrule{2-6}
    Anomaly & MSL & 55 & 40 & (44653, 11664, 73729) & \scalebox{0.8}{Spacecraft} \\
    \cmidrule{2-6}
    Detection & SMAP & 25 & 40 & (108146, 27037, 427617) & \scalebox{0.8}{Spacecraft} \\
    \cmidrule{2-6}
     & SWaT & 51 & 40 & (396000, 99000, 449919) & \scalebox{0.8}{Infrastructure} \\
    \cmidrule{2-6}
    & PSM & 25 & 40 & (105984, 26497, 87841)& \scalebox{0.8}{Server Machine} \\
    \bottomrule
    \end{tabular}
    \end{scriptsize}
    \end{sc}
  \end{threeparttable}
  \vspace{-5pt}
\end{table}

\subsubsection{Classification}
The time series classification task is a typical understanding task, which requires the model to make a holistic judgment based on the global representation. It is worth mentioning that, different from the multivariate time series prediction task that has been discussed more, part of the variate information and variate correlation in the classification of time series samples may be indispensable for accurate classification, so the multivariate modeling ability of the model is tested. In this experiment, we choose the multivariate data set UEA Archive~\cite{Bagnall2018TheUM} and the univariate UCR dataset~\cite{dau2019ucr} together as the benchmark data set. For a fair comparison, we keep our evaluation aligned with Time-Series-Library~\cite{wang2024tssurvey}. See Table~\ref{tab:downstream_dataset} for detailed information on the dataset.

\subsubsection{Anomaly Detection}
Multivariate time series anomaly detection tasks normally inherit the characteristics of having various anomalous segment lengths. Some anomalous segments are within 10 data points while others may be up to hundreds of data points. To study a representation that is capable of detecting subtle anomalies, we shrink the patch length to 4 during the large scale pertaining. Correspondingly, we also reduce the hidden dimension to 256 and the attention layers to 4 to avoid over-fitting. During downstream fine-tuning, we adopt a different input length of 40 compared to previous works such as TimesNet~\cite{wu2022timesnet}, corresponding to 10 input tokens. For a fair comparison, we keep our evaluation process aligned with previous works implemented in the Time Series Library. See Table~\ref{tab:downstream_dataset} for detailed details of the dataset.

\subsubsection{Imputation}

The multivariate time series imputation task we employ requires the model to impute against continuous missing values that occur frequently in real scenarios. At the same time, our test uses the multivariate benchmark data set to test the ability of the model to use the mutual hint between variates for collaborative imputation. Our benchmark datasets include (1) ETT~\cite{zhou2021informer} containing 7 variates of power transformers, (2) ECL~\cite{wu2021autoformer} consisting of hourly electricity consumption data from 321 customers (3) Weather~\cite{wu2021autoformer} consisting of 21 meteorological variates. For a fair comparison, we keep our evaluation process aligned with previous works implemented in Timer~\cite{liu2024timer}. See Table~\ref{tab:downstream_dataset} for details of the dataset.

\subsubsection{Short-term Forecasting}

Compared with the long-term forecasting of time series, the short-term forecasting of time series has a higher ability demand for the model to capture the global trend and overall pattern of the data in the lookback window, which adapts to the multi-granularity structure capture ability of TimesBERT. The experiment is carried out on the M4~\cite{M4team2018dataset}, which includes six univariate sequences with different change frequencies from various real scenes and can effectively prove the ability of the model to perform on data with different change patterns. For a fair comparison, we keep our evaluation aligned with previous works implemented in the Time-Series-Library. See Table~\ref{tab:downstream_dataset} for details of the dataset.

\section{Representation Analysis}
\begin{figure}[htbp]
\begin{center}
    \centerline{\includegraphics[width=1\columnwidth]{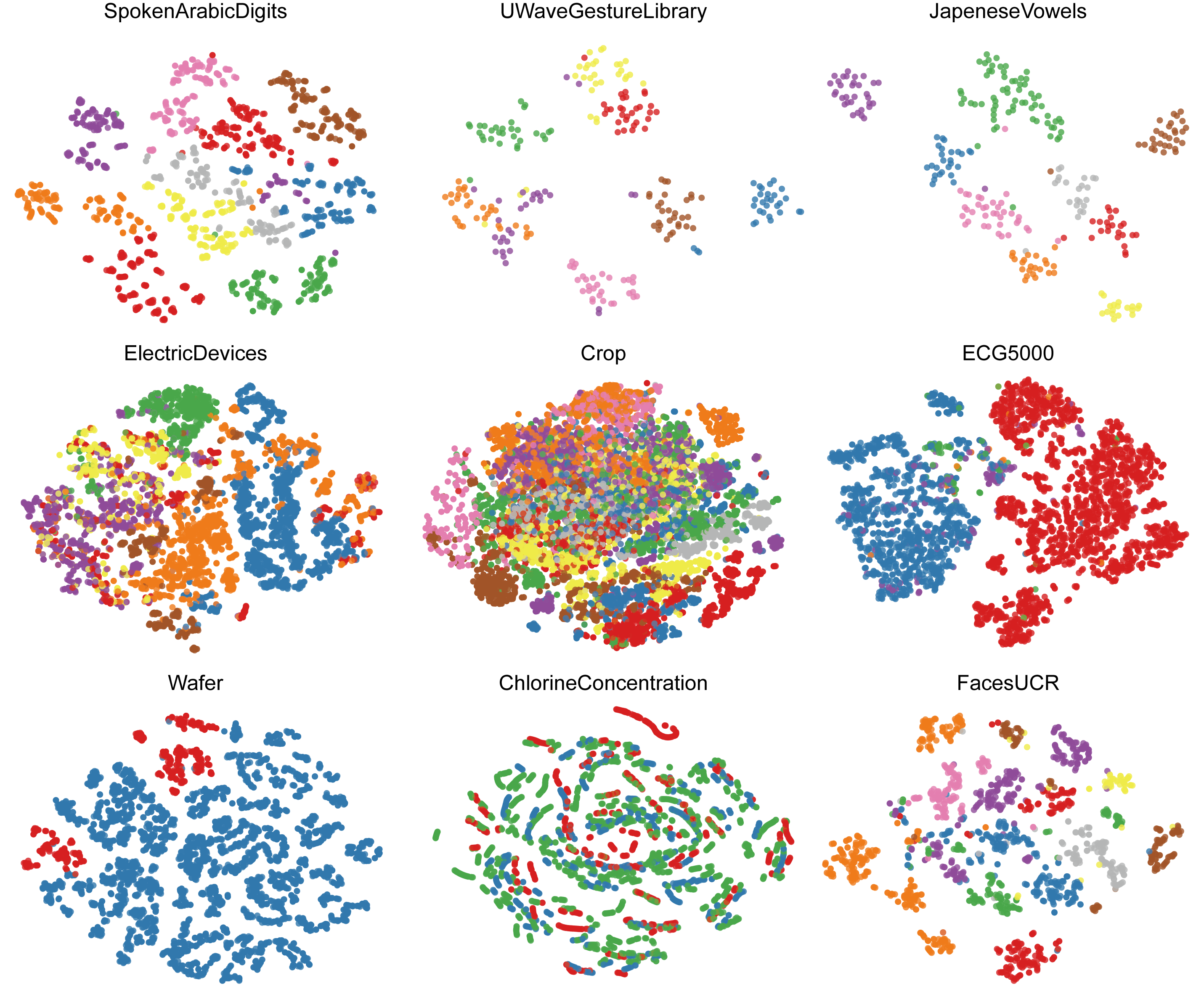}}
    \vspace{0pt}
	\caption{The t-SNE representation visualization of pre-trained TimesBERT on examples from the UEA and UCR classification datasets. Each color in visualization stands for a specific target class in an example dataset.} 
	\label{fig:representation}
\end{center}
\vspace{-10pt}
\end{figure}

To verify the effectiveness of TimesBERT representation, we perform 2D t-SNE~\cite{van2008visualizing} representation visualization on many classification datasets in a zero-shot setting. In Figure~\ref{fig:representation}, we present the example visualization of ``SpokenArabicDigits'',  ``UWaveGestureLibrary'', and ``JapeneseVowels'' from the UEA archive as well as ``ElectricDevices'', ``Crop'', ``ECG5000'', ``Wafer'', ``ChlorineConcentration'', and ``FacesUCR'' from the UCR archive. Small clusters of the same color in visualization suggest that samples from a specific class tend to have more similar high-dimensional representations. Such results show that TimesBERT is capable of acquiring distinct representations of different classes after systematic pertaining without data-specific fine-tuning.

\section{Full Results}
Due to the limitation of space in the main text, we present detailed results of all time series understanding tasks in this section, including time series classification, imputation, short-term forecasting, and anomaly detection. 

Time series classification results are as follows:  full results of UEA classification in Table~\ref{tab:classification_uea_full} and full results of UCR classification in Table~\ref{tab:classification_ucr_full}. Time series imputation results are as follows: imputation with 100\% samples in Table~\ref{tab:imputation_p100}, imputation with 20\% samples in Table~\ref{tab:imputation_p20}, imputation with 5\% samples in Table~\ref{tab:imputation_p5}, and full results of imputation in Table~\ref{tab:imputation_timesbert}. Full results of short-term forecasting in Table~\ref{tab:short_term_forecast_full}. Full results of anomaly detection in Table~\ref{tab:full_anomaly_results}.


\begin{table}[tbp]
  \caption{Full results for the classification task on UEA Archive. $\ast.$ in the Transformers indicates the name of $\ast$former. We report the classification accuracy (\%) as the result. The standard deviation is within 0.1\%. }\label{tab:classification_uea_full}
  \centering
  \vskip 0.1in
  \begin{sc}
  \begin{scriptsize}
  \renewcommand{\multirowsetup}{\centering}
  \setlength{\tabcolsep}{1.5pt}
\begin{tabular}{l|ccccccccccccccccccccccccc}
    \toprule
    \multirow{3}{*}{\scalebox{0.8}{Datasets / Models}} & \multicolumn{3}{c}{\scalebox{0.8}{Classical methods}} & \multicolumn{3}{c}{\scalebox{0.8}{RNN}}& \scalebox{0.8}{TCN} & \multicolumn{9}{c}{\scalebox{0.8}{Transformers}} & \multicolumn{2}{c}{\scalebox{0.8}{MLP}} & \multicolumn{2}{c}{\scalebox{0.8}{CNN}} & \multicolumn{2}{c}{\scalebox{0.8}{Pre-trained}}\\
    \cmidrule(lr){2-4}\cmidrule(lr){5-7}\cmidrule(lr){8-8}\cmidrule(lr){9-17}\cmidrule(lr){18-19}\cmidrule(lr){20-21}\cmidrule(lr){22-23}
    & \scalebox{0.6}{DTW} & \scalebox{0.6}{XGBoost} & \scalebox{0.6}{Rocket}  & \scalebox{0.6}{LSTM} & \scalebox{0.6}{LSTNet} & \scalebox{0.6}{LSSL} & \scalebox{0.6}{TCN} & \scalebox{0.6}{Trans.} & \scalebox{0.6}{Re.} & \scalebox{0.6}{In.} & \scalebox{0.6}{Pyra.} & \scalebox{0.6}{Auto.} & \scalebox{0.6}{Station.} &  \scalebox{0.6}{FED.} & \scalebox{0.6}{ETS.} & \scalebox{0.6}{Flow.} & \scalebox{0.6}{DLinear} & \scalebox{0.6}{LightTS.} &  \scalebox{0.6}{TimesNet}&  \scalebox{0.6}{MTCN} &    \scalebox{0.6}{MOMENT} &\scalebox{0.6}{\textbf{TimesBERT}} \\
\bottomrule
  \end{tabular}
    \end{scriptsize}
    \end{sc}
  \vspace{-10pt}
\end{table}

\begin{table*}[ht]
  \caption{Full results of the classification task on UCR Archive~\cite{dau2019ucr}.}
  \vskip 0.1in
  \label{tab:classification_ucr_full}
  \footnotesize
  \begin{small}
  \begin{sc}
  \linespread{2}
  \renewcommand{\multirowsetup}{\centering}
  \setlength{\tabcolsep}{0.8pt}
  \renewcommand\arraystretch{0.65}
  \begin{tabular}{l|ccccccccccccccccccccc}
    \toprule
\scalebox{0.8}{Datasets} &\scalebox{0.5}{DTW} &\scalebox{0.5}{TS2Vec} &\scalebox{0.5}{T-Loss} &\scalebox{0.5}{TNC} &\scalebox{0.5}{TS-TCC} &\scalebox{0.5}{TST} &\scalebox{0.5}{CNN} &\scalebox{0.5}{Encoder} &\scalebox{0.5}{FCN} &\scalebox{0.5}{MCDNN} &\scalebox{0.5}{ResNet} &\scalebox{0.5}{TimesNet} &\scalebox{0.5}{MTCN} &\scalebox{0.5}{t-LeNet} &\scalebox{0.5}{TWIESN} &\scalebox{0.5}{MLP} &\scalebox{0.5}{DLinear} &\scalebox{0.5}{PatchTST} &\scalebox{0.5}{MOMENT}&\scalebox{0.5}{GPT4TS}  &\scalebox{0.5}{\textbf{TimesBERT}}\\
    \toprule
\scalebox{0.6}{AllGestureWiimoteX} &\scalebox{0.6}{71.6} &\scalebox{0.6}{77.7} &\scalebox{0.6}{76.3} &\scalebox{0.6}{69.7} &\scalebox{0.6}{69.7} &\scalebox{0.6}{25.9} &\scalebox{0.6}{41.1} &\scalebox{0.6}{47.5} &\scalebox{0.6}{71.3} &\scalebox{0.6}{26.1} &\scalebox{0.6}{74.1} &\scalebox{0.6}{52.7} &\scalebox{0.6}{49.1} &\scalebox{0.6}{10.0} &\scalebox{0.6}{52.2} &\scalebox{0.6}{47.7} &\scalebox{0.6}{31.7} &\scalebox{0.6}{55.6} &\scalebox{0.6}{71.7} &\scalebox{0.6}{23.7} &\scalebox{0.6}{64.7}\\
\scalebox{0.6}{AllGestureWiimoteY} &\scalebox{0.6}{72.9} &\scalebox{0.6}{79.3} &\scalebox{0.6}{72.6} &\scalebox{0.6}{74.1} &\scalebox{0.6}{74.1} &\scalebox{0.6}{42.3} &\scalebox{0.6}{47.9} &\scalebox{0.6}{50.9} &\scalebox{0.6}{78.4} &\scalebox{0.6}{42.0} &\scalebox{0.6}{79.4} &\scalebox{0.6}{56.3} &\scalebox{0.6}{50.1} &\scalebox{0.6}{10.0} &\scalebox{0.6}{60.0} &\scalebox{0.6}{57.1} &\scalebox{0.6}{34.1} &\scalebox{0.6}{62.3} &\scalebox{0.6}{76.7} &\scalebox{0.6}{16.0} &\scalebox{0.6}{72.1}\\
\bottomrule
  \end{tabular}
  \end{sc}
  \end{small}
\end{table*}

\begin{table*}[ht]
  \caption{Imputation with $100\%$ samples. Pre-training benefit $\Delta\%$ is calculated as the ratio of decreased imputing error in MSE.}
  \vskip 0.1in
  \label{tab:imputation_p100}
  \footnotesize
  \begin{small}
  \begin{sc}
  \linespread{2}
  \renewcommand{\multirowsetup}{\centering}
  \setlength{\tabcolsep}{6.8pt}
  \renewcommand\arraystretch{1.4}
  \begin{tabular}{l|ccc|ccc|ccc|ccc}
    \toprule
    Mask Ratio & \multicolumn{3}{c|}{12.5\%} & \multicolumn{3}{c|}{25.0\%} & \multicolumn{3}{c|}{37.5\%} & \multicolumn{3}{c}{50.0\%} \\
    \cmidrule(lr){1-1} \cmidrule(lr){2-4} \cmidrule(lr){5-7} \cmidrule(lr){8-10} \cmidrule(lr){11-13}
    Pre-trained & None & 260B & $\Delta\%$ & None & 260B & $\Delta$\% & None & 260B & $\Delta\%$ & None & 260B & $\Delta\%$ \\
    \toprule
ETTh1   & 0.097 & 0.084 & 13.4 & 0.112 & 0.103 & 8.0  & 0.140 & 0.125 & 10.7 & 0.173 & 0.157 & 9.2  \\
ETTh2   & 0.137 & 0.077 & 43.8 & 0.114 & 0.084 & 26.3 & 0.128 & 0.097 & 24.2 & 0.150 & 0.116 & 22.7 \\
ETTm1   & 0.076 & 0.054 & 28.9 & 0.080 & 0.059 & 26.3 & 0.101 & 0.078 & 22.8 & 0.125 & 0.105 & 16.0 \\
ETTm2   & 0.062 & 0.040 & 35.0 & 0.052 & 0.046 & 12.0 & 0.072 & 0.055 & 23.7 & 0.084 & 0.066 & 20.9 \\
ECL     & 0.086 & 0.081 & 5.8  & 0.094 & 0.088 & 6.4  & 0.104 & 0.098 & 5.8  & 0.118 & 0.110 & 6.8  \\
Weather & 0.120 & 0.050 & 58.1 & 0.072 & 0.055 & 23.9 & 0.079 & 0.062 & 22.0 & 0.080 & 0.068 & 15.5  \\
    
    \bottomrule
  \end{tabular}
  \end{sc}
  \end{small}
\end{table*}

\begin{table*}[ht]
  \caption{Imputation with $20\%$ samples. Pre-training benefit $\Delta\%$ is calculated as the ratio of decreased imputing error in MSE.}
  \vskip 0.1in
  \label{tab:imputation_p20}
  \footnotesize
  \begin{small}
  \begin{sc}
  \linespread{2}
  \renewcommand{\multirowsetup}{\centering}
  \setlength{\tabcolsep}{6.8pt}
  \renewcommand\arraystretch{1.4}
  \begin{tabular}{l|ccc|ccc|ccc|ccc}
    \toprule
    Mask Ratio & \multicolumn{3}{c|}{12.5\%} & \multicolumn{3}{c|}{25.0\%} & \multicolumn{3}{c|}{37.5\%} & \multicolumn{3}{c}{50.0\%} \\
    \cmidrule(lr){1-1} \cmidrule(lr){2-4} \cmidrule(lr){5-7} \cmidrule(lr){8-10} \cmidrule(lr){11-13}
    Pre-trained & None & 260B & $\Delta\%$ & None & 260B & $\Delta$\% & None & 260B & $\Delta\%$ & None & 260B & $\Delta\%$ \\
    \toprule
ETTh1   & 0.294 & 0.147 & 50.0 & 0.286 & 0.162 & 43.4 & 0.291 & 0.182 & 37.5 & 0.298 & 0.208 & 30.2 \\
ETTh2   & 0.228 & 0.094 & 58.8 & 0.234 & 0.102 & 56.4 & 0.237 & 0.119 & 49.8 & 0.250 & 0.138 & 44.8 \\
ETTm1   & 0.545 & 0.086 & 84.2 & 0.396 & 0.099 & 75.0 & 0.379 & 0.118 & 68.9 & 0.380 & 0.148 & 61.1 \\
ETTm2   & 0.143 & 0.049 & 65.7 & 0.127 & 0.056 & 55.9 & 0.125 & 0.065 & 48.0 & 0.132 & 0.080 & 39.4 \\
ECL & 0.100 & 0.088 & 12.0 & 0.109 & 0.095 & 12.8 & 0.121 & 0.105 & 13.2 & 0.135 & 0.119 & 11.9 \\
Weather & 0.162 & 0.067 & 58.6 & 0.160 & 0.073 & 54.4 & 0.160 & 0.083 & 48.1 & 0.148 & 0.090 & 39.2 \\
    
    \bottomrule
  \end{tabular}
  \end{sc}
  \end{small}
\end{table*}

\begin{table*}[ht]
  \caption{Imputation with $5\%$ samples. Pre-training benefit $\Delta\%$ is calculated as the ratio of decreased imputing error in MSE.}
  \vskip 0.1in
  \label{tab:imputation_p5}
  \footnotesize
  \begin{small}
  \begin{sc}
  \linespread{2}
  \renewcommand{\multirowsetup}{\centering}
  \setlength{\tabcolsep}{6.8pt}
  \renewcommand\arraystretch{1.4}
  \begin{tabular}{l|ccc|ccc|ccc|ccc}
    \toprule
    Mask Ratio & \multicolumn{3}{c|}{12.5\%} & \multicolumn{3}{c|}{25.0\%} & \multicolumn{3}{c|}{37.5\%} & \multicolumn{3}{c}{50.0\%} \\
    \cmidrule(lr){1-1} \cmidrule(lr){2-4} \cmidrule(lr){5-7} \cmidrule(lr){8-10} \cmidrule(lr){11-13}
    Pre-trained & None & 260B & $\Delta\%$ & None & 260B & $\Delta$\% & None & 260B & $\Delta\%$ & None & 260B & $\Delta\%$ \\
    \toprule
ETTh1   & 0.364 & 0.207 & 43.1 & 0.363 & 0.220 & 39.4 & 0.367 & 0.236 & 35.7 & 0.377 & 0.249 & 34.0 \\
ETTh2   & 0.229 & 0.102 & 55.5 & 0.237 & 0.117 & 50.6 & 0.240 & 0.134 & 44.2 & 0.253 & 0.154 & 39.1 \\
ETTm1   & 0.740 & 0.122 & 83.5 & 0.748 & 0.155 & 79.3 & 0.759 & 0.192 & 74.7 & 0.780 & 0.237 & 69.6 \\
ETTm2   & 0.169 & 0.054 & 68.0 & 0.171 & 0.066 & 61.4 & 0.175 & 0.077 & 56.0 & 0.182 & 0.092 & 49.5 \\
ECL     & 0.168 & 0.094 & 44.0 & 0.171 & 0.103 & 39.8 & 0.176 & 0.114 & 35.2 & 0.183 & 0.128 & 30.1 \\
Weather & 0.162 & 0.079 & 51.2 & 0.162 & 0.082 & 49.4 & 0.165 & 0.089 & 46.1 & 0.169 & 0.104 & 38.5  \\
    
    \bottomrule
  \end{tabular}
  \end{sc}
  \end{small}
\end{table*}

\begin{table*}[ht]
  \caption{Full results of the imputation task.}
  \vskip 0.1in
  \label{tab:imputation_timesbert}
  \footnotesize
  \begin{small}
  \begin{sc}
  \linespread{2}
  \renewcommand{\multirowsetup}{\centering}
  \setlength{\tabcolsep}{4.7pt}
  \renewcommand\arraystretch{1.4}
  \resizebox{\linewidth}{!}{
  \begin{tabular}{l|cccc|cccc|cccc}
    \toprule
    Model & \multicolumn{4}{c|}{PatchTST} & \multicolumn{4}{c|}{MOMENT} & \multicolumn{4}{c}{Timer}  \\
    \cmidrule(lr){1-1} \cmidrule(lr){2-5} \cmidrule(lr){6-9} \cmidrule(lr){10-13} 
    Mask Ratio & 12.5\% & 25.0\% & 37.5\% & 50.0\% & 12.5\% & 25.0\% & 37.5\% & 50.0\% & 12.5\% & 25.0\% & 37.5\% & 50.0\% \\
    \toprule
ETTh1   & 0.250 & 0.250 & 0.266 & 0.283 & 0.238 & 0.253 & 0.265 & 0.285 & 0.273 & 0.283 & 0.294 & 0.312 \\
ETTh2   & 0.122 & 0.131 & 0.149 & 0.162 & 0.128 & 0.145 & 0.157 & 0.179 & 0.177 & 0.186 & 0.195 & 0.209 \\
ETTm1   & 0.147 & 0.173 & 0.211 & 0.258 & 0.137 & 0.165 & 0.201 & 0.248 & 0.352 & 0.345 & 0.371 & 0.413 \\
ETTm2   & 0.068 & 0.076 & 0.086 & 0.099 & 0.077 & 0.085 & 0.097 & 0.114 & 0.161 & 0.171 & 0.176 & 0.158 \\
ECL     & 0.105 & 0.115 & 0.128 & 0.143 & 0.095 & 0.102 & 0.113 & 0.127 & 0.122 & 0.130 & 0.139 & 0.152 \\
Weather & 0.073 & 0.078 & 0.086 & 0.099  & 0.077 & 0.085 & 0.100 & 0.113 & 0.157 & 0.146 & 0.147 & 0.158  \\
\bottomrule
  \end{tabular}
  }
  \end{sc}
  \end{small}
\end{table*}

\begin{table*}[ht]
  \caption{Full results of the short-term forecasting task. We follow the same protocol as~\citet{wu2022timesnet}.}
  \label{tab:short_term_forecast_full}
  \vspace{-5pt}
  \vskip 0.15in
  \centering
  \begin{small}
  \begin{sc}
  \renewcommand{\multirowsetup}{\centering}
  \setlength{\tabcolsep}{1.8pt}
  \renewcommand\arraystretch{1.2}
  \resizebox{\linewidth}{!}{
  \begin{tabular}{l|c|cccccccccccccccccccc}
    \toprule
    \multicolumn{2}{l|}{\multirow{2}{*}{Method}} & \textbf{TimesBERT} & ModernTCN & TimesNet & iTrans. & Koopa & NHiTS & DLinear & PatchTST & MICN & TiDE & MOMENT & NBEATS  \\ 
    \multicolumn{2}{l|}{} & \textbf{(Ours)} & \citeyearpar{wu2022timesnet} & \citeyearpar{donghao2024moderntcn} & \citeyearpar{liu2023itransformer} & \citeyearpar{liu2023koopa} & \citeyearpar{challu2023nhits} & \citeyearpar{zeng2023transformers}  & \citeyearpar{nie2022time} &  \citeyearpar{wang2022micn} & \citeyearpar{das2023long} & \citeyearpar{goswami2024moment} & \citeyearpar{oreshkin2019n} \\
    \toprule
    \multirow{3}{*}{\rotatebox{90}{Year}}
     & SMAPE & 13.312 & 13.226 & 13.387 & 13.923 & 13.352 & 13.371 & 13.866 & 13.517 & 14.532 & 15.320 & 20.649 & 13.466 \\
     & MASE  & 2.986  & 2.957 & 2.996  & 3.214  & 2.997  & 3.025  & 3.006  & 3.031  & 3.359  & 3.540  & 4.757  & 3.059  \\
     & OWA   & 0.783  & 0.777 & 0.786  & 0.830  & 0.786  & 0.790  & 0.802  & 0.795  & 0.867  & 0.910  & 1.230  & 0.797  \\
    \midrule
    \multirow{3}{*}{\rotatebox{90}{Quarter}}
     & SMAPE & 10.004  & 9.971 & 10.100 & 10.757 & 10.159 & 10.454 & 10.689 & 10.847 & 11.395 & 11.830 & 10.849 & 10.074 \\
     & MASE  & 1.158  & 1.167 & 1.182  & 1.283  & 1.189  & 1.219  & 1.294  & 1.315  & 1.379  & 1.410  & 1.305  & 1.163  \\
     & OWA   & 0.876  & 0.878 & 0.890  & 0.956  & 0.895  & 0.919  & 0.957  & 0.972  & 1.020  & 1.050  & 0.968  & 0.881  \\
\bottomrule
  \end{tabular}
  }
  
  \end{sc}
  \end{small}
\end{table*}

\begin{table}[tbp]
  \caption{Full results of the anomaly detection task. The P, R, and F1 represent the precision, recall, and F1-score (\%) respectively. F1-score is the harmonic mean of precision and recall. A higher value of P, R, and F1 indicates a better performance.}\label{tab:full_anomaly_results}
  \vskip 0.05in
  \centering
  \renewcommand{\multirowsetup}{\centering}
  \setlength{\tabcolsep}{1.4pt}
  \begin{sc}
  \begin{tabular}{lc|ccc|ccc|ccc|ccc|ccc|c}
    \toprule
    \multicolumn{2}{c}{\scalebox{0.8}{Dataset}} & 
    \multicolumn{3}{c}{\scalebox{0.8}{\rotatebox{0}{SMD}}} &
    \multicolumn{3}{c}{\scalebox{0.8}{\rotatebox{0}{MSL}}} &
    \multicolumn{3}{c}{\scalebox{0.8}{\rotatebox{0}{SMAP}}} &
    \multicolumn{3}{c}{\scalebox{0.8}{\rotatebox{0}{SWaT}}} & 
    \multicolumn{3}{c}{\scalebox{0.8}{\rotatebox{0}{PSM}}} & \scalebox{0.8}{Avg. F1} \\
    \cmidrule(lr){3-5} \cmidrule(lr){6-8}\cmidrule(lr){9-11} \cmidrule(lr){12-14}\cmidrule(lr){15-17}
    \multicolumn{2}{c}{\scalebox{0.8}{Metric}} & \scalebox{0.8}{P} & \scalebox{0.8}{R} & \scalebox{0.8}{F1} & \scalebox{0.8}{P} & \scalebox{0.8}{R} & \scalebox{0.8}{F1} & \scalebox{0.8}{P} & \scalebox{0.8}{R} & \scalebox{0.8}{F1} & \scalebox{0.8}{P} & \scalebox{0.8}{R} & \scalebox{0.8}{F1} & \scalebox{0.8}{P} & \scalebox{0.8}{R} & \scalebox{0.8}{F1} & \scalebox{0.8}{(\%)}\\
    \toprule
        \scalebox{0.85}{LSTM} &
        \scalebox{0.85}{\citeyearpar{Hochreiter1997LongSM}} %
        & \scalebox{0.85}{78.52} & \scalebox{0.85}{65.47} & \scalebox{0.85}{71.41} 
        & \scalebox{0.85}{78.04} & \scalebox{0.85}{86.22} & \scalebox{0.85}{81.93}
        & \scalebox{0.85}{91.06} & \scalebox{0.85}{57.49} & \scalebox{0.85}{70.48} 
        & \scalebox{0.85}{78.06} & \scalebox{0.85}{91.72} & \scalebox{0.85}{84.34} 
        & \scalebox{0.85}{69.24} & \scalebox{0.85}{99.53} & \scalebox{0.85}{81.67}
        & \scalebox{0.85}{77.97} \\
\bottomrule
    \end{tabular}
    \end{sc}
  \vspace{-10pt}
\end{table}

\section{Showcases}

\begin{figure*}[ht]
\begin{center}
    \center{\includegraphics[width=\textwidth]{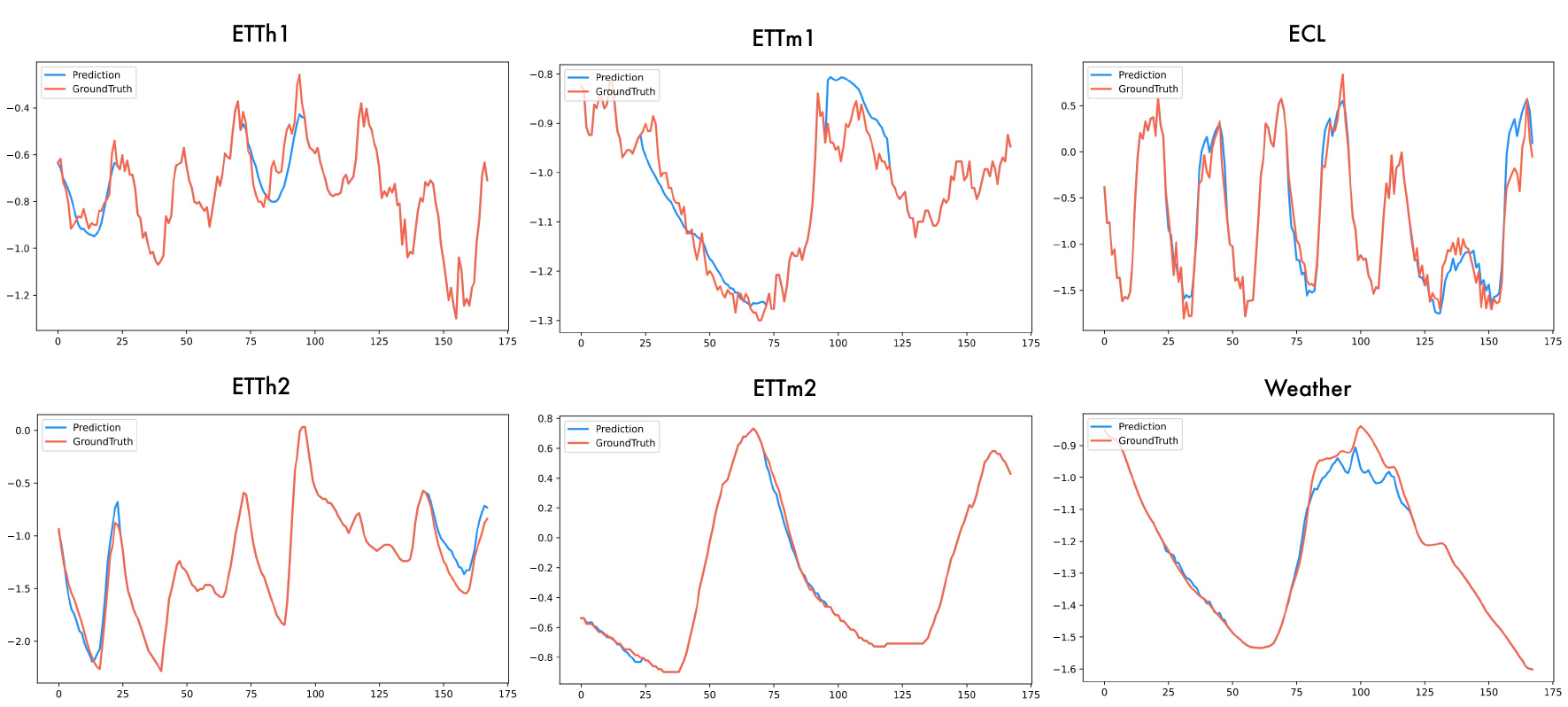}}
    \vspace{-20pt}
	\caption{Visualization of imputation results of TimesBERT on six benchmark datasets.}
	\label{fig:imputation_showcases}
\end{center}
\end{figure*}

\begin{figure*}[ht]
\begin{center}
    \center{\includegraphics[width=\textwidth]{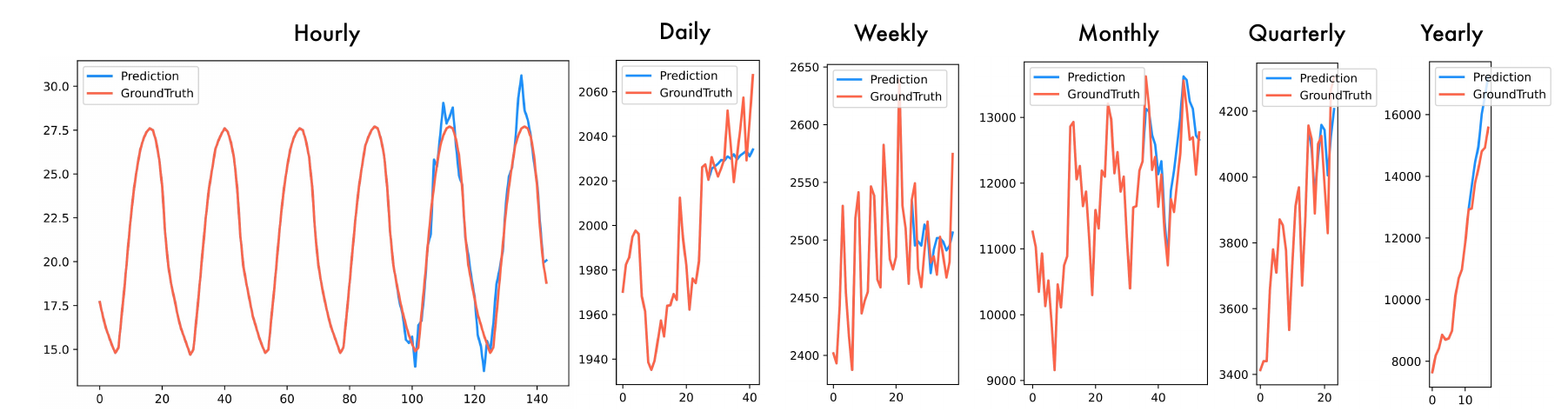}}
    \vspace{-20pt}
	\caption{Visualization of short-term forecasting results of TimesBERT on M4 dataset.}
	\label{fig:short_term_forecasting_showcases}
\end{center}
\end{figure*}

In this section, we provide the visualization results of TimesBERT on two downstream tasks of time series imputation and anomaly detection, corresponding to Figures~\ref{fig:imputation_showcases} and ~\ref{fig:short_term_forecasting_showcases}, respectively.

\end{document}